\documentclass[10pt,twocolumn,letterpaper]{article}

\usepackage[pagenumbers]{wacv}   

\newcommand{\algoname}{\mbox{\textsc{CoordFormer}}}
\newcommand{\semseg}{SS}

\definecolor{hlblue}{RGB}{220,235,255}
\definecolor{hlorange}{RGB}{255,235,210}
\definecolor{hlred}{RGB}{255,220,220}

\usepackage{soul}
\setuldepth{foobar}

\usepackage{multirow}
\usepackage{threeparttable}
 \usepackage{float} 
\usepackage{amsmath}
  \newcommand{\dsblock}[1]{\multicolumn{6}{c}{\textbf{#1}}}
\newcommand{\metrheader}{%
  $F_{\beta}^{x}\!\uparrow$ & $F_{\beta}^{w}\!\uparrow$ &
  $\mathcal{M}\!\downarrow$ & $S_m\!\uparrow$ &
  $E_{\phi}^{m}\!\uparrow$ & $\mathrm{HCE}_{\gamma}\!\downarrow$
}
\usepackage{pifont}
\newcommand{\cmark}{\ding{51}} 
\newcommand{\xmark}{\ding{55}} 

\usepackage{bbm}
\usepackage[table]{xcolor} 
\definecolor{sota}{RGB}{220,245,220}      
\definecolor{secondsota}{RGB}{255,245,204}
\newcommand{\best}[1]{\cellcolor{sota}\textbf{#1}}
\newcommand{\second}[1]{\underline{\cellcolor{secondsota}#1}}
\usepackage[ruled,vlined]{algorithm2e}
\usepackage{microtype}
\newcommand{\datakidney}{\mbox{\textsc{KPIs}}\xspace}
\newcommand{\sizetwo}[2]{\ensuremath{#1\!\times\!#2\xspace}}
\newcommand{\sizethree}[3]{\ensuremath{#1\mkern-2mu\times\mkern-2mu#2\mkern-2mu\times\mkern-2mu#3\xspace}}

\definecolor{wacvblue}{rgb}{0.21,0.49,0.74}
\usepackage[breaklinks,colorlinks,allcolors=wacvblue]{hyperref}
\usepackage{orcidlink}   

\def\confName{WACV}
\def\confYear{2027}

\title{\algoname{}: Give Me Any Coordinates and I Will Give You Labels}

\author{%
Iacopo Curti\textsuperscript{1}\,\orcidlink{0009-0005-4732-6440} \quad
Pierluigi Zama Ramirez\textsuperscript{2}\,\orcidlink{0000-0001-7734-5064} \quad
Alioscia Petrelli\textsuperscript{3}\,\orcidlink{0000-0000-0000-0000} \quad
Luigi Di Stefano\textsuperscript{1}\,\orcidlink{0000-0001-6014-6421}\\[0.3em]
\textsuperscript{1}CVLab, University of Bologna \qquad
\textsuperscript{2}Ca' Foscari University of Venice \qquad
\textsuperscript{3}SINA, company
}

\begin{document}
\twocolumn[{%
\renewcommand\twocolumn[1][]{#1}%
\maketitle
\begin{center}
    \centering
    \includegraphics[width=0.9\linewidth]{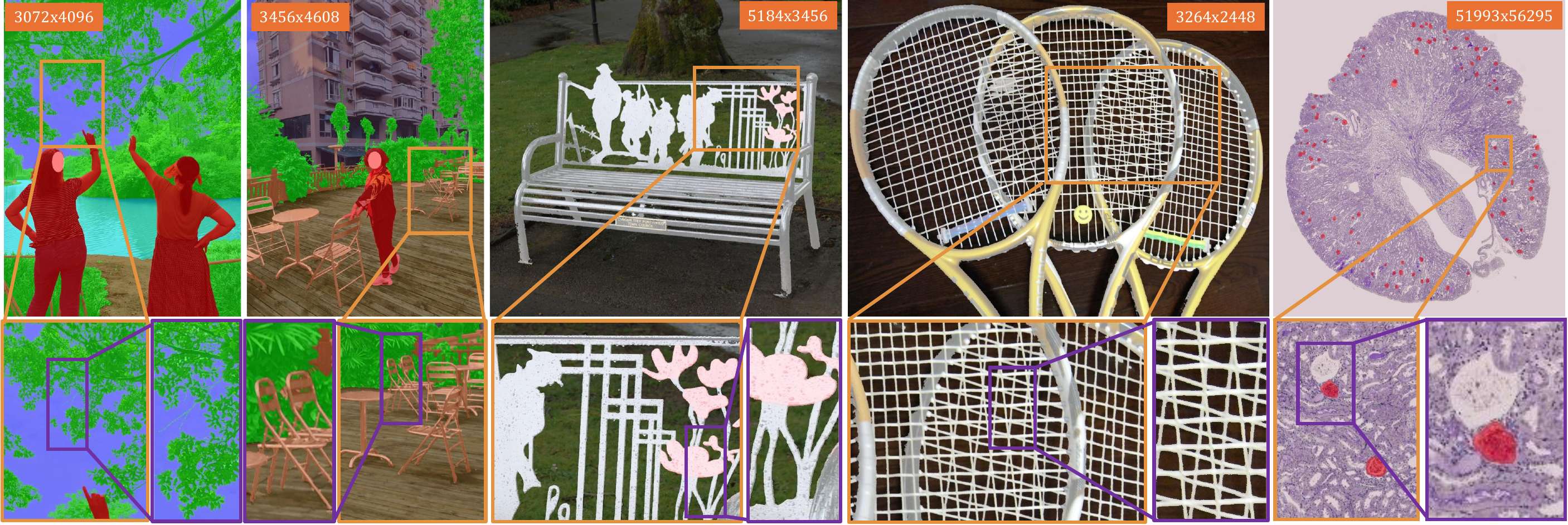}
    \captionof{figure}{\textbf{Segmentation results by \algoname{} on MaSS13K \cite{Xie2025mass13k} (left), DIS5K \cite{qin2022dis5k} (center) and \datakidney \cite{deng2025kpis} (right).} \algoname{} predicts labels directly at selected pixel locations, enabling highly detailed segmentation of very-high-resolution images. This design allows \algoname{} to capture thin structures and fine boundaries.}
    \label{fig:teaser}
\end{center}
}]

\begin{abstract}
Semantic segmentation on very-high-resolution images remains challenging due to the high computational cost and the difficulty of capturing fine-grained details. We propose \algoname{}, a novel coordinate-based architecture for semantic segmentation that predicts labels at arbitrary spatial locations through a Coordinate Decoder equipped with a Localized Cross-Attention mechanism. The decoder combines coordinate embeddings with high-resolution local patch features and interacts with global tokens extracted from a downsampled image processed by a ViT foundation encoder, enabling rich semantic context while preserving pixel-level precision.
This design enables flexible inference at arbitrary resolutions while keeping memory low on very-high-resolution inputs, and supports an efficient semantic-edge-focused strategy that concentrates computation along boundaries, maintaining fine-grained accuracy while reducing latency and computational cost.
\algoname{} achieves state-of-the-art performance on MaSS13K and outperforms comparably sized and higher-parameter methods on DIS5K and KPIs, demonstrating its effectiveness for high-quality, very-high-resolution semantic segmentation. 
\end{abstract}

\section{Introduction}
\label{sec:intro}
Semantic Segmentation (\semseg{}) is a core problem in computer vision that aims to assign a semantic label to every pixel in an image, enabling dense understanding of visual scenes. Recent years have witnessed remarkable progress in \semseg{}, largely driven by deep learning techniques  \cite{minaee2021survey}.
Most existing \semseg{} models, however, are designed and evaluated on standard-resolution datasets, typically around 1 megapixel  \cite{chen2018deeplabv3plus, xie2021segformer, cheng2022mask2former}.
Although several methods have explored high-resolution semantic segmentation  \cite{Lin2017RefineNet, wang2020hrnet, cavagnero2024pem}, they are usually developed based on benchmarks such as Cityscapes  \cite{cordts2016cityscapes} and ADE20K  \cite{zhou2017ade20k}, where average image resolutions range from 1 to 4 megapixels and annotated masks exhibit limited structural complexity.
Consequently, these approaches tend to overlook the challenges of very-high-resolution segmentation (\textgreater{} 4 Mpx) and often fail to capture thin structures or fine-grained details accurately, as highlighted in  \cite{Xie2025mass13k}.
However, achieving finely detailed segmentation on very-high-resolution images is critical for numerous real-world applications, including image editing, augmented reality, and  high-impact domains such as medical image analysis.
To systematically study the problem, MaSS13K~\cite{Xie2025mass13k} was introduced at CVPR 2025 as a benchmark specifically designed to reveal the limitations of existing models in fine-grained, very-high-resolution semantic segmentation. Along with it, MaSSFormer~\cite{Xie2025mass13k} was proposed as an initial attempt to address the task and serve as a strong baseline for future research.
Motivated by these findings, we take a step further and propose a new \semseg{} architecture, termed \algoname, capable of addressing the challenging scenario set forth by MaSS13K. At its core lies a Coordinate Decoder  \cite{park2019deepsdf, sitzmann2020siren} equipped with a tailored Localized Cross-Attention mechanism.
The decoder includes two complementary components: an MLP that processes spatial coordinates, and a ViT-style patch embedder that operates on small patches extracted from high-resolution images. Together, these modules generate coordinate-aware query tokens enriched with pixel-level, high-resolution details.
These query tokens interact, through the proposed Localized Cross-Attention, with global tokens derived from a downsampled version of the entire image, which is processed by a Vision Transformer foundation model such as DINOv3~ \cite{simeoni2025dinov3}. This interaction results in highly contextualized tokens, rich in semantic information while remaining aware of the fine-grained details present in the input image, enabling precise labeling at the queried spatial coordinates.\\
Our novel framework offers several advantages.
First, the coordinate-based strategy allows to predict labels only for a few selected pixels  during inference, such as a particular region of interest, or to flexibly set the output resolution according to the target application.
Consequently, our design allows explicit control over the computational cost at inference time, making it possible to choose the optimal trade-off between accuracy, efficiency, and memory constraints.
This property stands in stark contrast to previous methods such as MaSSFormer~\cite{Xie2025mass13k}, which require processing the entire input image and thus incur significant memory and computational overhead when handling very-high-resolution data. Indeed,  MaSSFormer resizes all images to a fixed 12.6 Mpx resolution during inference, highlighting its limited flexibility in dealing with varying image scales.
Furthermore, by leveraging the coordinate-based formulation, we introduce an efficient inference strategy. This approach focuses computation along semantic boundaries reducing inference time, while maintaining high accuracy on thin structures and fine details.
\noindent
Second, our coordinate-based decoding strategy, when coupled with a strong pre-trained foundation encoder, enables extremely precise predictions at the full very-high-resolution of the input images, as illustrated in \cref{fig:teaser}.
When evaluated on the challenging, finely detailed, very-high-resolution semantic segmentation task introduced by MaSS13K, \algoname{} achieves state-of-the-art performance, surpassing existing approaches by a substantial margin.
To further assess the precision of our predictions, we also evaluate \algoname{} on the Dichotomous Image Segmentation (DIS) benchmark, DIS5K~ \cite{qin2022dis5k}, where it outperforms both models of comparable size and higher-parameter ones, confirming its effectiveness in capturing fine-grained details. 
We additionally evaluate our method on the Kidney Pathology Image Segmentation benchmark (\datakidney \cite{deng2025kpis}), where it surpasses recent task-specific networks \cite{dominguezmantes2026muvit, tang2024holohisto} under the protocol \cite{dominguezmantes2026muvit}.
This result confirms that our method transfers to a markedly different segmentation domain, even on the gigapixel-scale whole-slide images that characterize \datakidney.
The code is publicly available on
\href{https://iacopo97.github.io/CoordFormer/}{GitHub}. \noindent
The main contributions of this paper are:
\begin{itemize}
\item A novel Coordinate Decoder for very-high-resolution semantic segmentation, capable of producing highly detailed, spatially precise predictions.
\item A new Localized Cross-Attention mechanism tailored to our coordinate-based framework, enabling efficient integration of local and global information.
\item An efficient inference strategy, enabled by our coordinate-based formulation, that concentrates computation along semantic boundaries, dramatically reducing latency and cost while preserving accuracy.
\item
We achieve state-of-the-art performance on MaSS13K and superior results on DIS5K and \datakidney, demonstrating the generality of our method
\end{itemize}

\section{Related Works}
\paragraph{\normalfont\textbf{Semantic Segmentation.}}
Early deep learning models tackled  the semantic segmentation task with Fully Convolutional Networks (FCNs) \cite{long2015fcn}, with notable examples including UNet \cite{ronneberger2015unet}, PSPNet \cite{zhao2017pspnet}, and DeepLab \cite{chen2018deeplab, chen2017deeplabv2, chen2017deeplabv3, chen2018deeplabv3plus}, among others \cite{csurka2022survey}. The introduction of transformers \cite{dosovitskiy2020vit} and attention mechanisms \cite{vaswani2017attention} has reshaped the field, enabling models like Swin Transformer \cite{liu2021swin}, DPT \cite{ranftl2021dpt}, SegFormer \cite{xie2021segformer}, MaskFormer \cite{cheng2021maskformer}, and SegNext \cite{guo2022segnext} to capture long-range dependencies and multi-scale context. Recent universal segmentation models such as Mask2Former \cite{cheng2022mask2former} further exemplify the impact of attention-based designs.
A critical component of these models is the decoder, which reconstructs dense predictions from encoded features. UNet \cite{ronneberger2015unet} introduced skip connections; DeepLabv3+ \cite{chen2018deeplabv3plus} used atrous convolution; SegFormer employed a lightweight MLP decoder; Mask2Former combined pixel and transformer decoders. In this work, we propose a novel Coordinate Decoder with Localized Cross-Attention that enables finely-detailed semantic predictions with controlled latency and memory consumption.
\paragraph{\normalfont\textbf{High-resolution Semantic Segmentation.}}
Handling high-resolution images remains a major challenge in semantic segmentation due to high computational and memory demands. Several methods have been proposed to address this task, either by explicitly targeting high-resolution processing \cite{Lin2017RefineNet, chen2019collaborative, wang2020hrnet, shan2021uhrsnet, gu2022multi, yuan2021hrformer, qi2022entityseg, Ke2023samHQ, cavagnero2024pem, liu2024sparse}, or by focusing on general efficiency \cite{takikawa2019gated, howard2019mobilenetv3, zhao2018icnet, shim2023feedformer, yu2021bisenet, wan2025seaformer++, xu2023pidnet, ni2024cgrseg}, thereby enabling high-resolution inference under standard hardware constraints.
\begin{figure*}[t]
    \centering
    \includegraphics[width=0.78\linewidth]{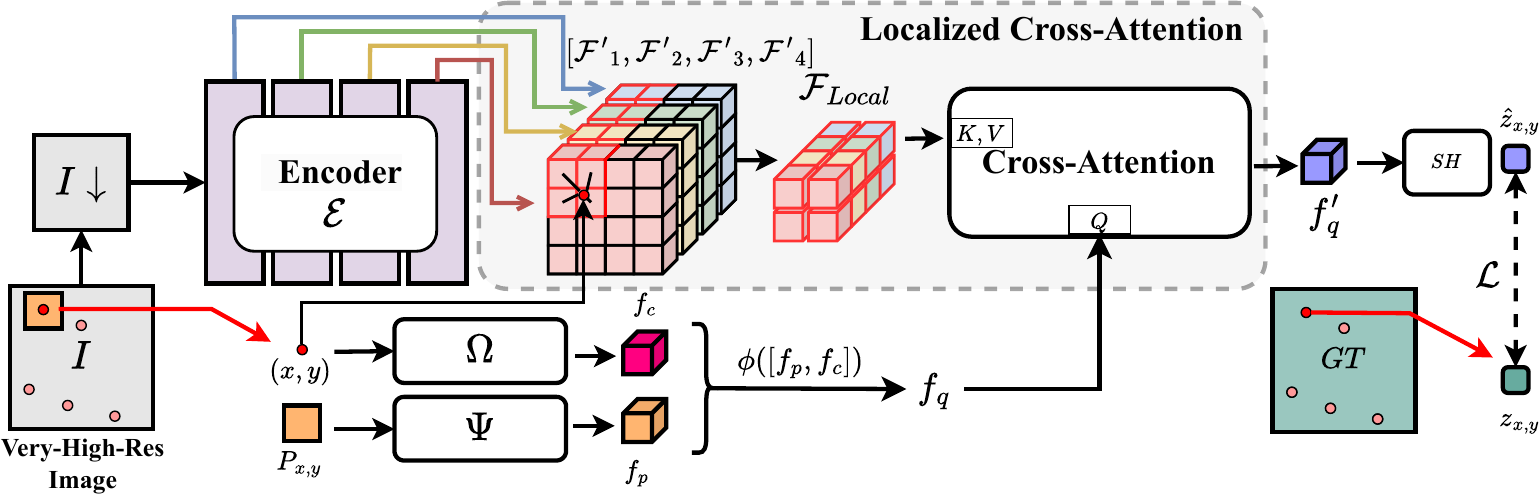}
    \caption{\textbf{\algoname{} Architecture and Training Overview.}
    Given a very-high-resolution image ($>4$Mpx), a ViT-based foundation encoder, $\mathcal{E}$, extracts globally contextualized tokens from a downsampled input, while a Coordinate Decoder builds coordinate-aware query features, $f_q$, from local high-resolution patches, $P_{x,y}$, and pixel coordinates, $(x,y)$. These queries interact with nearby global tokens through Localized Cross-Attention obtaining features, $f_q'$ rich in both semantics and fine-details. These are passed to a Segmentation Head (SH) to predict class logits, $\hat{z}_{x,y}$, at the queried locations. During training, pixel and label information is sampled from random coordinates of images and ground truths, and the network is optimized with multi-class Cross-entropy, per-class Dice and per-class Cross-entropy losses.}
    \label{fig:pipeline}
\end{figure*}
Beyond resource efficiency, 
high-resolution segmentation demands precise delineation of fine structures. 
Refinement strategies \cite{cheng2020cascadepsp, yuan2020segfix, shen2022high, kirillov2020pointrend, zhao2024airm} and edge-aware techniques \cite{li2020edge, wang2022edge} can improve output sharpness, yet many approaches still underperform in very-high-resolution settings -- partly due to limited resolution of most dataset (e.g., 2-4 Mpx) and the lack of finely-detailed annotations. Recent, coordinate-based refiners \cite{kirillov2020pointrend, zhao2024airm} operate on coordinates, but require a dense segmentor and correct its output; differently, our method predicts labels directly at arbitrary coordinates.
To foster research on  very-high resolution image understanding, a recent semantic segmentation benchmark has been proposed, MaSS13K \cite{Xie2025mass13k}, together with a task-specific baseline, MaSSFormer. 
Other segmentation tasks in which boundary precision is essential are Image Matting and
  Dichotomous Image Segmentation (DIS).
Methods for these tasks \cite{qin2022dis5k, zhou2023fpdis, pei2023udun, zheng2024birefnet} are typically precise on boundaries, yet do not focus on capturing high-level semantic concepts, as the objective is limited to foreground-background separation without semantic classification.
Regarding very-high-resolution segmentation, the medical community introduced \datakidney \cite{deng2025kpis} at MICCAI 2024. It is a histopathology dataset for glomeruli segmentation in gigapixel-scale images. Although recent methods \cite{dominguezmantes2026muvit, tang2024holohisto} address this task, they are incapable of boundary-based inference, remaining fully dense.
%
In this work, we introduce a novel framework capable of producing detailed predictions on very high-resolution inputs for both semantic and dichotomous image segmentation.
\paragraph{\normalfont\textbf{Attention Mechanisms.}}
Attention mechanisms have gained traction in computer vision following their deployment in Transformers \cite{vaswani2017attention}. Vision Transformer (ViT) applied self-attention to image patches for global context modeling \cite{dosovitskiy2020vit}, while DETR employed cross-attention to align object queries with image features \cite{carion2020detr}. Deformable DETR improved efficiency via sparse, content-adaptive attention \cite{zhu2020deformable}, and Axial-DeepLab factorized attention along spatial axes to reduce complexity \cite{wang2020axial}. Swin Transformer introduced window-based attention with shifted windows for hierarchical representations \cite{liu2021swin}, and MaskFormer proposed masked attention to focus decoding within predicted regions \cite{cheng2021maskformer}. Recently, SegNeXt revisited convolutional attention, fusing efficiency and spatial priors with attention-driven modeling \cite{guo2022segnext}. In this work, we propose a Localized Cross-Attention layer specifically designed for our coordinate decoder.
\section{Method}
We address the problem of \textit{very-high-resolution semantic segmentation}, where the goal is to assign a semantic label to every pixel of a high-resolution input image.
Formally, given an image $ I \in \mathbb{R}^{H \times W \times 3}$, the objective is to learn a function, 
$f_\theta: \mathbb{R}^{H \times W \times 3} \rightarrow \mathbb{R}^{H \times W \times K}$, that assigns a probability distribution over the label set of $K$ classes, to each pixel $(x, y) \in I$.
Unlike conventional approaches that produce dense predictions for all pixels simultaneously, our method, named \algoname{}, adopts a \textit{coordinate-based} formulation that allows querying arbitrary spatial locations at inference time.
Specifically, given an input coordinate
\begin{equation}
(x, y) \in [0, W - 1] \times [0, H - 1],
\end{equation}
our model predicts the semantic logit as
\mbox{$
\hat{z}_{x,y} = f_\theta(I, (x, y)) \in  \mathbb{R}^{K}.
$}
This formulation provides flexible control over output resolution, computational cost, latency and memory consumption, while enabling precise segmentation results. An overview of the architecture and training of \algoname{} is shown in \cref{fig:pipeline}.
\subsection{Architecture}\paragraph{\normalfont\textbf{Encoder.}} The encoder is responsible for extracting globally-contextualized semantic representations from the input image. 
Given the very-high resolution of the data, directly processing the image at its native scale would be computationally prohibitive. 
To address this, the input image $I$ is first downsampled to a manageable resolution $I^\downarrow$ of size $H^\downarrow \times W^\downarrow$ ($H^\downarrow =W^\downarrow \in \{2048, 2044\}$ in our experiments), which is then processed by a powerful Vision Transformer (ViT) foundation model:
\begin{equation}
\{\mathcal{F}_i\}_{i=1}^{4} = \mathcal{E}(I^\downarrow)
\end{equation}
where $\mathcal{E}$ denotes the encoder (e.g., DINOv3~\cite{simeoni2025dinov3}), and $\{\mathcal{F}_i\}_{i=1}^{4}$ are feature maps of dimensions $h \times w \times c_e$ from four different transformer blocks.
Here, $h = H^\downarrow / p_e$ and $w = W^\downarrow / p_e$ correspond to the spatial resolution of the feature maps given the encoder patch size $p_e$, and $c_e$ denotes the number of feature channels. Each global feature map $\mathcal{F}_i$ is projected to a lower dimensional channel space, $c_{l}=c_e / 4$, obtaining $\mathcal{F'}_i$.
For each spatial position, the tokens from these layers are concatenated to a single, rich representation:
\begin{equation}
\mathcal{F}_g = [\mathcal{F'}_1, \mathcal{F'}_2, \mathcal{F'}_3, \mathcal{F'}_4] \in \mathbb{R}^{h \times w \times 4c_l}.
\end{equation}
In the feature map, $\mathcal{F}_g$, each token encodes semantic information associated with a specific image location at multiple levels of abstraction. 
These globally-contextualized tokens provide a context-aware representation of the scene, serving as the semantic foundation for the subsequent coordinate-based decoding stage.
\paragraph{\normalfont\textbf{Coordinate Decoder.}}
The Coordinate Decoder predicts the semantic logit at a specific spatial location of the original high-resolution image $I$ by building a coordinate-aware query from the input coordinate and its local high-resolution patch. This query token is subsequently used in the \textit{Localized Cross-Attention} (LCA) module to interact with the global tokens produced by the encoder, effectively combining fine local information with global semantic context.
Formally, given an input query coordinate $(x, y)$, we extract a small image patch $P_{x,y}$, of size $p \times p$ ($p = 8$ in our experiments), whose top-left corner is aligned with this position, from the high-resolution image $I$. 
This patch is processed by a lightweight MLP patch embedder, $\Psi$, to produce a compact feature vector encoding local texture and boundary details:
\begin{equation}
f_p = \Psi(P_{x,y}), \quad \Psi: \mathbb{R}^{3p^2} \rightarrow \mathbb{R}^{d_p},
\end{equation}
In parallel, the 2D coordinate $(x, y)$ is normalized to the range $[0,1]$ and transformed to Fourier-feature embeddings  \cite{tancik2020fourier} by $\Omega$,
that maps the spatial position into a frequency embedding space:
\begin{equation}
f_c = \Omega(x, y), \quad \Omega: \mathbb{R}^{2} \rightarrow \mathbb{R}^{d_c},
\end{equation}
The resulting patch feature and coordinate embedding are concatenated and projected through a sequence of linear layers with ReLU, $\phi$ to form a \textit{coordinate-aware query token}:
\begin{equation}
f_q = \phi([f_p, f_c]), \quad \phi: \mathbb{R}^{d_p + d_c} \rightarrow \mathbb{R}^{c_q},
\end{equation}
which encodes both pixel-level high-frequency details and precise spatial awareness.
\paragraph{\normalfont\textbf{Localized Cross-Attention.}}
The LCA module performs a cross-attention operation that allows the coordinate-aware query token $f_q$ to interact with a small subset of encoder representations corresponding to its spatial neighborhood, so as to keep computational cost low. 
Given the query coordinate $(x, y)$ in the high-resolution image space, we first map it to the coordinate system of the encoder feature map, which has been downsampled both by the input resizing and by the encoder patching operation. 
Let $s_H = H / H^\downarrow$ and $s_W = W / W^\downarrow$ denote the spatial scaling factors between the original image and the encoder input, along the vertical and horizontal axes respectively, and let $p_e$ denote the encoder patch size, the corresponding location $(x^\downarrow, y^\downarrow)$ in the feature map is computed as
\begin{equation}
(x^\downarrow, y^\downarrow) = \left( \frac{x}{s_W \cdot p_e}, \frac{y}{s_H \cdot p_e} \right),
\end{equation}
which accounts for both the input downsampling and the encoder’s patch resolution.
%
%
We project the input coordinate $(x,y)$ onto the downsampled token grid $\mathcal{F}_g$, obtaining
$(x^\downarrow, y^\downarrow)$.
We then extract a local neighborhood of $k$ tokens around $(x^\downarrow, y^\downarrow)$ via bilinear
sampling on $\mathcal{F}_g$ (implemented with \texttt{grid\_sample}), yielding a local set
$\mathcal{F}_{\text{local}}$ 
of size $k$.
In our implementation we sample $k=4$ local features
corresponding to a $2 \times 2$ neighborhood around $(x^\downarrow, y^\downarrow)$.
A Cross-Attention (CA) operation is then applied between the coordinate-aware query token $f_q$ and the sampled local encoder tokens $\mathcal{F}_{\text{local}}$ (serving as keys and values):
\begin{equation}
f_q' = \mathrm{CA}(f_q, \mathcal{F}_{\text{local}})
\end{equation}
where $f_q'$ denotes the refined query token enriched with global semantic context.
This LCA module drastically reduces computational complexity, which becomes linear in $k$ rather than in the total number of global tokens. 
Despite the restriction to a small neighborhood, we observe no loss in segmentation quality, as the encoder tokens are highly contextualized and rich in semantic information. 
\paragraph{\normalfont\textbf{Segmentation Head.}}
The refined query feature $f_q'$ produced by the LCA module is passed through a lightweight \textit{Segmentation Head} (SH) to obtain the class logits associated with the queried coordinate. 
The SH is implemented as a small MLP with a Batch-Norm layer that maps the contextualized feature into the semantic space of $K$ classes:
\begin{equation}
\hat{z}_{x,y} = \mathrm{SH}(f_q'), \quad \mathrm{SH}: \mathbb{R}^{c_q} \rightarrow \mathbb{R}^{K},
\end{equation}
where $\hat{z}_{x,y}$ denotes the predicted logits for coordinate $(x,y)$.
\subsection{Training}
During training, we supervise the model by randomly sampling coordinates $(x, y)$ from the high-resolution image domain of size $H \times W$. 
For each sampled coordinate, the corresponding local patch is extracted, processed by the encoder--decoder architecture, and passed through SH to obtain the class logits $\hat{z}_{x,y} \in \mathbb{R}^{K}$.
The prediction at each queried coordinate is supervised with a weighted combination of the standard Multi-class Cross-Entropy (CE) loss , $K$ class-specific Binary Dice (D) 
and Binary Cross-Entropy (BCE) losses computed over the individual logit maps.
The overall loss is defined as
\begin{equation}
\mathcal{L}=\lambda_{\mathrm{CE}}\mathcal{L}_{\mathrm{CE}}
+ \tfrac{1}{K}\!\sum_{k=1}^{K}(\lambda_{\mathrm{D}}\mathcal{L}_{\mathrm{D}}^{k}
+ \lambda_{\mathrm{BCE}}\mathcal{L}_{\mathrm{BCE}}^{k})
\label{eq:loss}
\end{equation}
where 
$\mathcal{L}_{\mathrm{CE}}=L_{\mathrm{CE}}(\hat{z}_{x,y}, z_{x,y})$, $\mathcal{L}_{\mathrm{D}}^{k}=L_{\mathrm{D}}(\hat{z}^k_{x,y}, z^k_{x,y})$, $\mathcal{L}_{\mathrm{BCE}}^{k}=L_{\mathrm{BCE}}(\hat{z}^k_{x,y}, z^k_{x,y})$
, $z_{x,y}$ denote the ground truth class at location $(x,y)$, and $\lambda_{\mathrm{CE}}$, $\lambda_{\mathrm{D}}$, and  $\lambda_{\mathrm{BCE}}$ are scalar weights that balance the loss terms.
This formulation encourages both accurate pixel-level classification through the Cross-Entropy loss and sharp, well-defined segmentation boundaries through the Dice loss component.
The training optimizes encoder, decoder, and SH parameters using AdamW optimizer with learning rate $1e^{-4}$. The model has been trained on 4 A100 for approximately 170 epochs (with early stopping), using a batch size of 16 MaSS13K images \cite{Xie2025mass13k} and sampling 16{,}384 random coordinates per image.
\begin{figure}[t]
    \centering\includegraphics[width=0.98\linewidth]{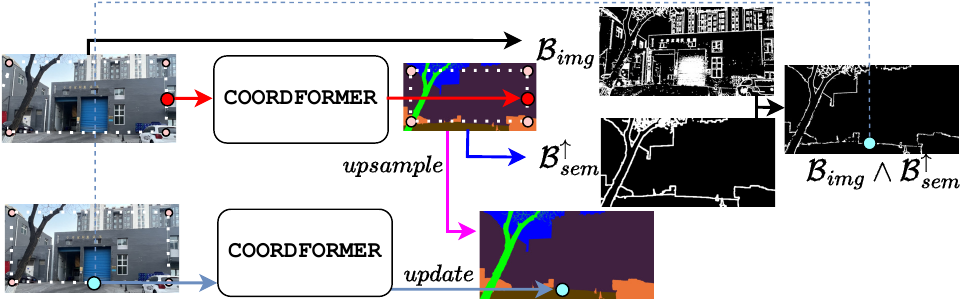}
  \centering
  \caption{\textbf{Semantic-Edge-Focused Inference Overview} }
  \label{fig:sefi}
\end{figure}
\subsection{Inference}
\paragraph{\textbf{Coordinate-based Inference.}}
Our framework supports flexible inference by predicting labels only at queried spatial coordinates (\cref{fig:pipeline}) combined with corresponding high resolution patch, rather than producing a dense segmentation map in a single forward pass. 
Given a set of query coordinates $\{(x_i, y_i)\}_{i=1}^{N}$, each coordinate and the corresponding patch are independently processed by \algoname{} to obtain its corresponding class logits $\hat{z}_i$ and consequently the label as $\textit{argmax}(\hat{z}_i)$.
This design offers control over the trade-off between accuracy, efficiency, and memory usage.
%
Dense predictions can be obtained by querying all pixels, while sparse ones can be produced by sampling only a subset of coordinates, allowing the model to adapt its computational load to different applications and hardware constraints.
\paragraph{\textbf{Semantic-Edge-Focused Inference.}}
To further improve efficiency, we introduce a \textit{Semantic-Edge-Focused Inference} (SEFI) strategy, shown in \cref{fig:sefi}, that concentrates computation along thin semantic boundaries. The procedure operates as follows. First, we sample coordinates with a fixed stride $s$ (set to $8$ in our experiments) and predict a coarse segmentation map.
From this coarse prediction, we extract a binary semantic boundary map, $\mathcal{B}_{\text{sem}}$, 
where a pixel is assigned value $1$ if it has at least one adjacent pixel which is predicted with a different semantic label. 
To increase robustness to segmentation errors, we also dilate the map by a square kernel of size $r$.
Since these boundaries are coarse, we also compute a binary image boundary map, $\mathcal{B}_{\text{img}}$, by applying a Sobel filter and thresholding to the RGB image, $I$, obtaining edges that better fits structural details but include both semantic and non-semantic contours. 
To retain only structural semantic boundaries, we perform a pixel-wise logical \texttt{AND} between the upsampled semantic boundary map, $\mathcal{B}_{\text{sem}}^{\uparrow}$ (resized to $H \times W$), and the Sobel boundary map:
\begin{equation}
\mathcal{B} = \mathcal{B}_{\text{sem}}^{\uparrow} \land \mathcal{B}_{\text{img}}.
\end{equation}
The resulting binary mask, $\mathcal{B}$, better highlights semantic boundaries. 
To obtain the full resolution prediction, we first upsample the coarse segmentation map, then compute predictions at the coordinates corresponding to the refined semantic boundaries. Finally, we update values at boundary locations in the coarse segmentation mask.
%
This strategy, enabled by our coordinate-based design, substantially reduces the pixels, required to achieve sharp and accurate dense segmentations. Additional inference protocols (e.g., region-of-interest querying) are described in the supplementary material (Sec. A4).
\section{Experimental Results}
\subsection{Very-High-Resolution Segmentation}
\begin{figure}[t]
\centering
\includegraphics[width=1\linewidth]{
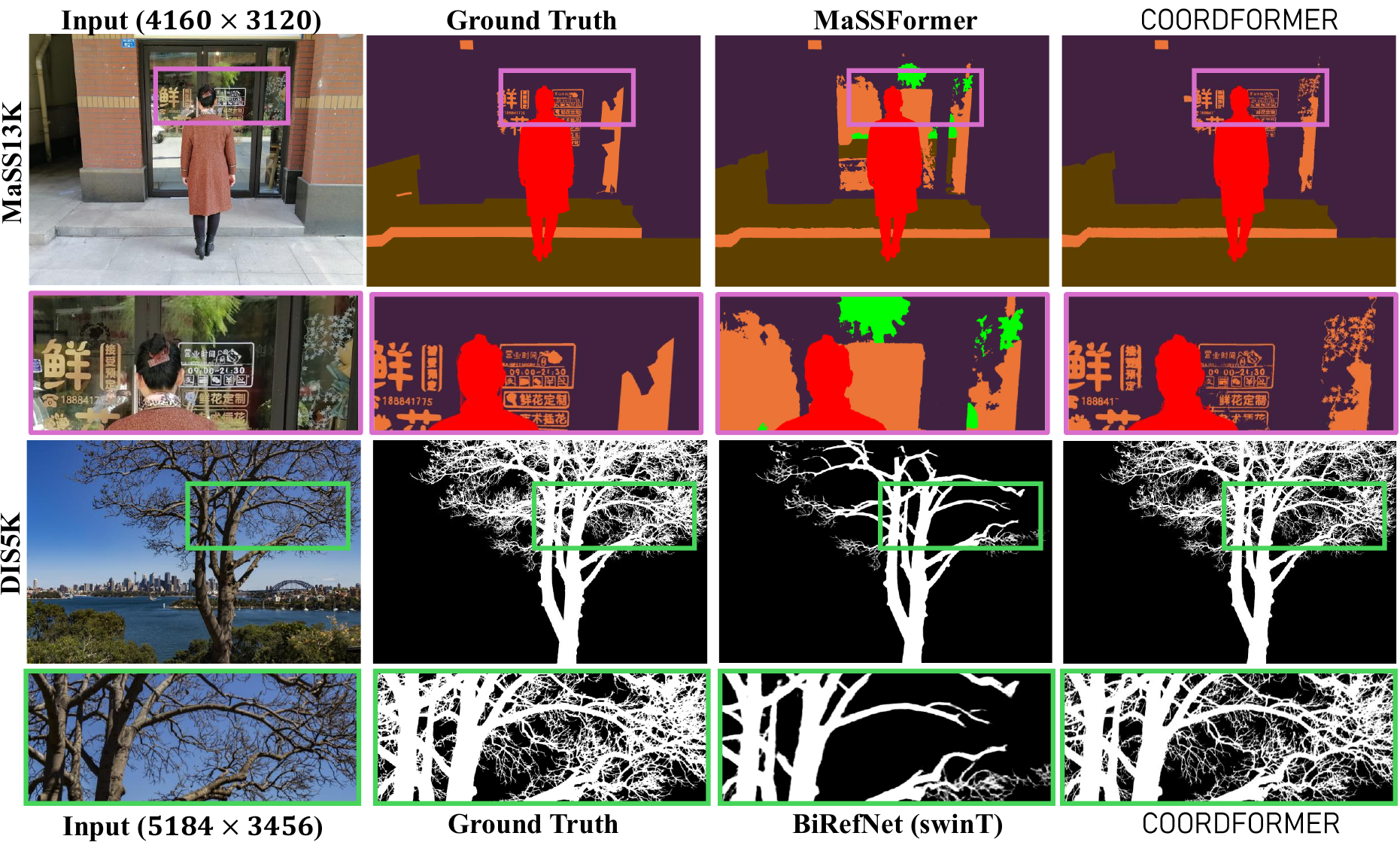}
\caption{\textbf{Qualitative comparisons to main competitors on MaSS13K (top) and DIS5K (Bottom).}}
\label{fig:qualitatives}
\end{figure}

\paragraph{
\textbf{Results on MaSS13K.}
}
We evaluate our method on MaSS13K~\cite{Xie2025mass13k}, a very-high-resolution benchmark for high-quality semantic segmentation, and report results in \cref{tab:mass_sota}. MaSS13K contains 13,348 real-world images, most at 12Mpx resolution,
split into 11,348/500/1,500 train/validation/test samples. The dataset provides high-quality masks for seven categories (human, vegetation, ground, sky, water, building, and others) and is characterized by exceptionally high annotation complexity - on average \(20\!\times\)–\(50\!\times\) higher than standard segmentation benchmarks. The benchmark primarily assesses general segmentation performance using mIoU, and the fine-grained precision of predictions using BIoU, and BF1 scores. As shown in \cref{tab:mass_sota}, our method with SEFI achieves state-of-the-art performance on both the validation and test sets, demonstrating a strong ability to recover fine-grained details in very-high-resolution scenarios. 
Improvements are particularly pronounced on boundary-focused metrics such as BIoU and BF1, which are most indicative of performance on thin structures and detailed edges. We emphasize that \algoname{} achieves remarkable gains in boundary metrics with fewer parameters than MaSSFormer, further demonstrating the effectiveness of the proposed coordinate-based architecture. 
Qualitative results in \cref{fig:teaser} and \cref{fig:qualitatives} highlight the effectiveness of our method in preserving complex structural details.
\begin{table}[t]
\centering
\begin{minipage}[t]{0.39\textwidth}
  \centering
   \resizebox{\linewidth}{!}{
  \begin{tabular}{lc|ccc|ccc|cc}
\toprule
& & \multicolumn{3}{c|}{MaSS-val (500)} & \multicolumn{3}{c|}{MaSS-test (1{,}500)} & \multicolumn{2}{c}{Model Stat.} \\
\textbf{Methods} & Backbone & mIoU$\uparrow$ & BIoU$\uparrow$ & BF1$\uparrow$ & mIoU$\uparrow$ & BIoU$\uparrow$ & BF1$\uparrow$ & Param & MACs \\
\midrule
BiSeNetv2~\cite{yu2021bisenet}   & --        & 71.55 & 25.05 & .3182 & 72.92 & 24.48 & .3171 & 3.35M  & 591 G \\
SegNeXt~\cite{guo2022segnext}       & MSCAN-B  & 87.71 & 39.93 & .4615 & 88.11 & 39.45 & .4596 & 27.57M & 1536 G \\
PIDNet-L~\cite{xu2023pidnet}       & --        & 82.28 & 31.30 & .3475 & 81.77 & 30.70 & .3479 & 37.08M & 1653 G \\
FeedFormer~\cite{shim2023feedformer} & lvt       & 87.17 & 42.06 & .4838 & 86.56 & 41.07 & .4789 & 4.65M & 300 G \\
SeaFormer-L~\cite{wan2025seaformer++} & --        & 86.78 & 38.61 & .4498 & 87.36 & 38.28 & .4489 & 13.95M & 303 G \\
\midrule
DeepLabv3+~\cite{chen2018deeplabv3plus} & R50  & 86.66 & 40.08 & .4718 & 85.14 & 38.65 & .4678 & 41.22M & 8008 G \\
UPerNet~\cite{xiao2018unified}          & R50  & 82.03 & 35.85 & .4181 & 81.98 & 35.61 & .4170 & 64.04M & 11373 G \\
\midrule
MaskFormer~\cite{cheng2021maskformer}    & R50  & 83.27 & 38.61 & .4399 & 83.22 & 37.90 & .4393 & 41.31M & 2396 G \\
Mask2Former~\cite{cheng2022mask2former}  & R50  & 88.28 & 47.40 & .5458 & 88.00 & 46.13 & .5330 & 44.01M & 3123 G \\
MPFormer~\cite{zhang2023mp}        & R50 & 87.76 & 47.81 & .5513 & 87.18 & 47.17 & .5486 & 43.9M & 4155 G \\
PEM~\cite{cavagnero2024pem}                  & R50 & 83.41 & 40.51 & .4675 & 83.38 & 39.99 & .4644 & 35.5M & 1859 G \\
\midrule
MaSSFormer-Lite \cite{Xie2025mass13k} & R18 & 87.11 & 45.35 & .5137 & 86.13 & 43.28 & .5086 & 15.07M & 771 G \\
MaSSFormer \cite{Xie2025mass13k} & R50 & \second{88.97} & \second{48.97} & \second{.5639} & \second{88.21} & \second{48.39} & \second{.5593} & 37.42M & 2036 G \\
\midrule
\algoname{} & DINOv3\texttt{-}S+ & \best{\textbf{92.53}} & \best{\textbf{53.24}} & \best{\textbf{.5965}}     &\best{\textbf{92.30}}   &\best{\textbf{52.40}}  & \best{\textbf{.5916}} & 29.92M & 6430* G \\
\bottomrule
\end{tabular}}
\subcaption{\textbf{MaSS13K Results.}
  \label{tab:mass_sota}
}
\end{minipage}
\begin{minipage}[t]{0.39\textwidth}
  \centering

  \centering
  \resizebox{\linewidth}{!}{%
  \setlength{\tabcolsep}{4pt}
  \renewcommand{\arraystretch}{1.15}

  \begin{tabular}{l *{12}{c}|c}
    \toprule
    \multirow{2}{*}{\textbf{Methods}} 
     & \dsblock{DIS-TE (1–4) (2,000)} & \dsblock{DIS-VD (470)} & \multirow{2}{*}{\textbf{Params}}\\
    \cmidrule(lr){2-7} \cmidrule(lr){8-13} 
    & \metrheader  & \metrheader & (M)\\
    \midrule
    
    PSPNet \cite{zhao2017pspnet} &        .710   &     .620 &   .095   &  .755    &  .819    &  1442    & .691     &    .603  &         .102  &   .744   &    .802  &    1588 &- \\
    DeepLabv3+ \cite{chen2018deeplabv3plus}             & .678      &   .584    &  .105        &  .729     &   .810   &1365  &    .660 &  .568    & .114     &  .716    &.796      & 1520    &41 \\
    HRNet \cite{wang2020hrnet}            & .743      &   .658   &  .087    &   .781   &   .840   &   1432   &   .726   &      .641 &   .095    &     .767   &   .824   &    1560 &-   \\
    ICNet \cite{zhao2018icnet}            & .711     &    .622  &    .095  &     .758 &  .825    &   1359   &   .697   &     .609  &     .102  &   .747    &   .811   & 1503  &-   \\
    MBV3 \cite{howard2019mobilenetv3}            &  .729    &  .658    &   .085   &   .770   &  .850    &    1457  &     .714 &   .642   &    .092  &   .758   &    .841  &   1625  &- \\
    STDC2 \cite{fan2021bisenet_rethinking}            & .710 &.628 &.094 & .754  &   .832   &   1426       &   .696   &  .613    &  .103    &   .740   & .817     &   1598   &- \\
    \midrule
    IS-Net~\cite{qin2022dis5k}             &.799 &.726 &.070   &       .819    &    .858 &  1016   &  .791    & .717     &  .074    &  .813    &     .856 &  1116   &44 \\
    FP-DIS~\cite{zhou2023fpdis}
     &   .831   &  .770    &   .047  &   .847   &    .895  &    1165  &   .823   &  .763    &   .062   &   .843   &   .891   & 1309     & -\\
    UDUN~\cite{pei2023udun}            &  .831    &   .772   &  .057    &  .844    &     .892 &     977 &  .823    & .763     &  .059    &  .838    &  .892    &  1097  &25  \\
    BiRefNet\textsubscript{swinT}~\cite{zheng2024birefnet}        &.866 &.822 &.045 &.877 &.916 &980  &  .862    &  .819    &  .045    & .874     & .917     &  1070  &39  \\
    BiRefNet\textsubscript{swinB}~\cite{zheng2024birefnet}    &.891 &.855 &.036 &.898 &.933 &954      & .881      &  .844     & .039      &   .890    & .925 & 1029     & 101\\
    BiRefNet\textsubscript{swinL}~\cite{zheng2024birefnet}    &\second{.896} &\second{.858} &\second{.035} &\best{\textbf{.901}} &\second{.934} &\second{916}      & \second{.891}      &  \second{.854}     & \second{.038}      &   \second{.898}    & \second{.931} & \second{989}     & 215\\
    \midrule
    \algoname{}   & \best{\textbf{.899}}   &   \best{\textbf{.876}}  &  \best{\textbf{.032}}    & \second{.900}    &   \best{\textbf{.940}}   &    \best{\textbf{806}}     & \best{\textbf{.898}}   &   \best{\textbf{.879}}  &  \best{\textbf{.032}}    & \best{\textbf{.902}}     &   \best{\textbf{.942}}   &    \best{\textbf{787}}    &30 \\
    \bottomrule
  \end{tabular}}
  \subcaption{\textbf{DIS5K Results}.
    \label{tab:dis5k}
}
  

\end{minipage}
\begin{minipage}[t]{0.39\textwidth}
  \centering
  \resizebox{0.45\linewidth}{!}{
  \begin{tabular}{@{}lrr}
  \toprule
    \textbf{Methods} & Input size & DSC$\uparrow$ \\
    \midrule
    U-Net$^{**}$ \cite{ronneberger2015unet} & $\sizetwo{512}{512}$ & 0.5499\\
    SAM-ViT-H$^{**}$ \cite{kirillov2023segment} & $\sizetwo{1024}{1024}$ & 0.7724 \\
    HoloHisto-4K$^{**}$ \cite{tang2024holohisto}& $\sizetwo{3840}{2160}$ & 0.8454 \\
    MuViT$_{[1,8]}${\small +UNETR}$^{**}$ & $\sizethree{2}{512}{512}$ & \second{0.8958}\\
    \midrule
    \algoname{} & $\sizetwo{2048}{2048}$ & \best{\textbf{0.9178}}\\
    \bottomrule
  \end{tabular}}
    \subcaption{ \textbf{\datakidney{} (test set) Results.}
      \label{tab:kpis_sota}
}
\end{minipage}
\caption{\textbf{Very-High-Resolution Segmentation Results on MaSS13K \cite{Xie2025mass13k} (a) DIS5K \cite{qin2022dis5k} (b) and \datakidney (c) }. “$\uparrow$” (“$\downarrow$”) means higher (lower) is better. *MACs averaged over the dataset. ** indicates the score was taken from \cite{dominguezmantes2026muvit}. \colorbox{sota}{\strut \textbf{Best}}, \colorbox{secondsota}{\strut \underline{Second-Best}} } 
\end{table}
\paragraph{
\textbf{Results on DIS5K.}
}
We also evaluate \algoname{} on DIS5K~\cite{qin2022dis5k}, a benchmark for high-precision Dichotomous Image Segmentation (train/val/test: 3{,}000/470/2{,}000), and report results in \cref{tab:dis5k}. 
Despite not being tailored to the DIS task, \algoname{} achieves superior performance on nearly all metrics---except $S_m$ in the test set relative to BiRefNet with \textit{swinL}
---outperforming more parameter-intensive methods and clearly surpasses approaches of similar size.
In particular, compared to BiRefNet (\textit{swinT}) our model gets significant improvements across all the metrics, highlighting the generality of our coordinate-based design outside the semantic segmentation setting.
\paragraph{\textbf{Results on \datakidney.}}
On the \datakidney{} challenge (Task 2, WSI-level diseased glomeruli segmentation), which features gigapixel images (train/val/test: 30/8/12), \algoname{} achieves 0.9178 DSC, surpassing MuViT~\cite{dominguezmantes2026muvit} (Tab.~\ref{tab:kpis_sota}) under the same inference protocol. This shows that our method extends to the gigapixel medical domain, confirming its applicability across diverse very-high-resolution scenarios.
\subsection{Computational Analysis}\label{subsec:comput_ana}
\begin{figure}[t]
  \centering
    \begin{minipage}{\linewidth}
      \centering
      \begin{subfigure}[t]{0.67\linewidth}
        \centering
        \includegraphics[width=1\linewidth]{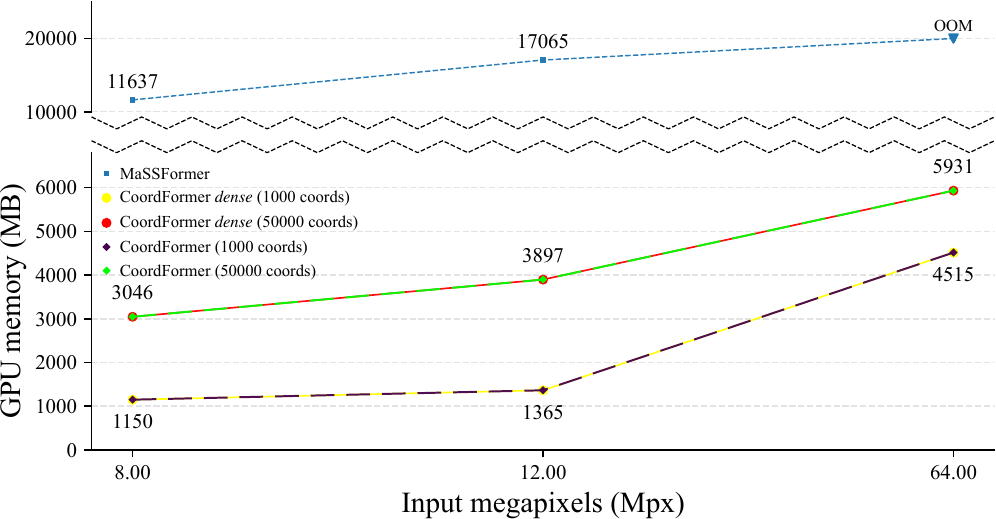}
        \subcaption{GPU memory vs input Mpx}
        \label{fig:table_mem_plots}
      \end{subfigure}\hfill
      \begin{subfigure}[t]{0.68\linewidth}
        \centering
        \includegraphics[width=1\linewidth]{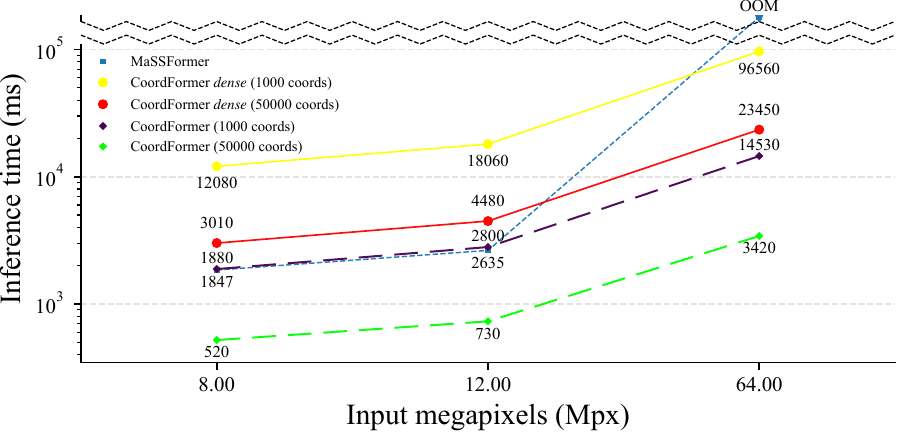}
        \subcaption{Inference time vs input Mpx}
        \label{fig:table_inf_plots}
      \end{subfigure}
    \end{minipage}%
  \caption{\textbf{GPU Memory footprint (a) and Inference Times (b) of \algoname{}, \algoname{} dense (no SEFI) and MaSSFormer}. GPU memory footprint (a) and inference time (b) are evaluated at input resolutions of 8, 12, and 64 Mpx, with \algoname{} tested using coordinate batch of 1k and 50k.}
  \label{fig:table_inf_mem_plots}
\end{figure}

\paragraph{\textbf{\algoname{} Inference time and Memory footprint.}}
The computational cost of \algoname{} (i.e., with SEFI), depends on the number of semantic-edge pixels processed at inference. 
Hence, in \cref{tab:mass_sota} we report MACs averaged over the images of MaSS13K. Note that the MaSS13K paper~\cite{Xie2025mass13k} contains a typo: the reported values are MACs (as counted by \texttt{fvcore}\footnote{\href{https://github.com/facebookresearch/fvcore}{https://github.com/facebookresearch/fvcore}}), not FLOPs, since fused multiply--accumulate is counted as a single operation. On average, our method uses more MACs than competing methods.
However, unlike standard dense decoders, \algoname{} allows explicit control over memory occupancy and latency: the number of processed coordinates can be freely adjusted to match the desired trade-off between accuracy, latency, and hardware constraints. In memory- or compute-constrained platforms, inference can be performed in small coordinate batches (down to pixel-by-pixel), substantially reducing GPU memory occupancy, albeit at the cost of increased inference time due to reduced parallelism. Conversely, in time-constrained applications, processing of large coordinate batches can be parallelized across cores and devices to minimize latency. Peculiarly, \algoname{} can compute an extremely high resolution output without the need to store a dense full-resolution feature map, since it does not require processing the full-resolution image in a single forward pass.
In order to assess their practical computational costs, we compare \algoname{} and MaSSFormer in terms of GPU memory footprint (left)
and inference time (right), in \cref{fig:table_inf_mem_plots}.
For our method, we consider \algoname{}, i.e. the default configuration with SEFI, and \algoname{} \textit{dense}, where all the coordinates within the image are processed (i.e., without SEFI). With both configurations, we evaluate multiple inference settings by varying the number of coordinates processed per batch (1k and 50k).  All experiments are performed on a single NVIDIA A6000 GPU with 48\,GB, and at three resolutions: 8, 12, and 64\,Mpx.
\noindent In terms of memory, \algoname{} scales gracefully with image resolution, thanks to its coordinate-based inference strategy.
Contrarily, MaSSFormer requires resizing images to $4096 \times 3072$ to handle very high-resolution inputs. \algoname{} and \algoname{} \textit{dense} show no difference in memory requirements, since during inference we still need to keep only a single coordinate batch (1k, 50k) in VRAM (the red and green curves overlap, as do the yellow and purple ones).
MaSSFormer requires way more memory than our models, and its official  implementation  yields an out-of-memory (OOM) error at 64\,Mpx. 
\algoname{} and \algoname{} \textit{dense}, instead, can successfully handle all input resolutions in the considered datasets, without any OOM, and with memory consumption primarily determined by the coordinate batch size.
Regarding inference time, \algoname{} is faster than MaSSFormer with the largest coordinate batches (50k), and comparable with 1k coordinate batches. In particular, \algoname{} takes  $\sim$$\frac{1}{3}$ of MaSSFormer's inference time up to 12\,Mpx (e.g., 520 vs.\ 1847\,ms with 8\,Mpx, 730 vs. 2635 ms with 12 Mpx) and, unlike MaSSFormer, can process also 64\,Mpx inputs. 
\algoname{} \textit{dense} is slower, yet, despite the way higher MACs, thanks to the parallelization, the runtime remains in the same order of magnitude as MaSSFormer (e.g., 4480 vs.\ 2635\,ms, CoordFormer \textit{dense} vs.\ MaSSFormer at 12\,Mpx).
Importantly, \algoname{} \textit{dense} achieves the best performance overall (see row~3 of \cref{tab:ablation}).
Further increasing the coordinate batch size on the hardware specified above does not yield additional gains, as CUDA core saturation prevents further parallelism.
%
Overall, these results show that MACs alone may not provide a clear picture when comparing the practical computational efficiency of different methods: although our method entails more MACs than competing approaches, its coordinate-based formulation enables effective parallelization and avoids common bottlenecks (e.g., memory-bandwidth limits, kernel-launch overheads, and poor parallelization), resulting in lower latency and smaller memory footprint in practice, achieving fast inference ($>$ 1\,fps up to 12\,Mpx) with affordable memory usage on consumer-level GPUs.
\subsection{Ablation studies}
\label{subsec:Ablation}
\paragraph{\textbf{Main Contributions Ablations.}}
\Cref{tab:ablation} reports the main contributions of \algoname{}. 
In the first row, we evaluate a baseline semantic segmentation model obtained by equipping DINOv3~\cite{simeoni2025dinov3} with the simple Segmentation Head (SH). This model processes images resized to $2048 \times 2048$, as \algoname{}, but the results on MaSS13K reveal that while it effectively captures semantic information, it struggles to produce precise segmentations.
Specifically, it achieves high mIoU of 90.09 but low BIoU of 28.72, highlighting limited boundary accuracy. If we compare the first row with all other rows, that include the Coordinate-Decoder, we note a substantial improvement across all metrics, particularly in boundary scores (BIoU and BF1), confirming the benefits of the coordinate-based design for very-high-resolution high-quality segmentation.
\begin{table}[t]
\centering
\resizebox{\linewidth}{!}{
\begin{tabular}{l|c|cc|ccc|cccccc}
\toprule
& Decoder & \multicolumn{2}{c|}{Inference} & \multicolumn{3}{c|}{MaSS-val (500)} & Model & Total & Encoder & Decoder & Decoder & Coord\\
\textbf{Methods} & Attention & $\mathcal{B}_{\text{sem}}$  & $\mathcal{B}_{\text{img}}$ & mIoU$\uparrow$ & BIoU$\uparrow$ & BF1$\uparrow$  & Params & MACs & MACs & MACs & MACs (1 coord) & Usage \% \\
\midrule

DINOv3 \cite{simeoni2025dinov3} + SH & - & \xmark &\xmark & 90.09 & 28.72 &  .3507 & 28.70M & 2960 G & 2950 G &  10 G & - & - \\
\algoname{} & CA &\xmark &\xmark & 92.17 & 52.82 & .5889  & 29.92M& 168270 G & 2950 G & 165320 G & 4.847 G & 100 \\
\algoname{} & LCA &\xmark &\xmark  & \best{\textbf{92.62}} & \best{\textbf{54.51}} & \best{\textbf{.6074}} &  29.92M& 26990 G&  2950 G& 24040 G & .002 G  & 100 \\
\algoname{} & CA  &\cmark &\cmark &92.12 & 51.87 & .5817  & 29.92M& 27160 G & 2950 G & 24210 G & 4.847 G & 14 \\
\algoname{} & LCA &\cmark &\xmark & \second{92.59} &\second{53.92} &\second{.6043} &29.92M &7360 G &2950 G & 4410 G & .002 G &18 \\
\midrule
\algoname{} & LCA &\cmark &\cmark & 92.53 & 53.24 & .5965 &  29.92M& 6430 G & 2950 G & 3480 G & .002 G & 14\\

\bottomrule
\end{tabular}}
\caption{\textbf{Contributions ablations on MaSS13K\cite{Xie2025mass13k}.} SH: Segmentation Head. CA: Cross-Attention. LCA: Localized Cross-Attention. “$\uparrow$”  means higher is better.  \colorbox{sota}{\strut \textbf{Best}}, \colorbox{secondsota}{\strut \underline{Second-Best}}}
\label{tab:ablation}
\end{table}
While MACs may be an inaccurate proxy for computational efficiency across different methods—which can exhibit varying degrees of inherent parallelism—they remain meaningful within these ablations, where we vary only the number of attended tokens/coordinates within our own method.
Comparing the performance of the second and third row, we can appreciate the benefits on efficiency and boundary metrics when using LCA rather than a simple Cross-Attention.
LCA requires far fewer MACs than standard CA when processing a single coordinate (0.002 G vs. 4.847 G) and substantially lower total MACs over the full image, while achieving higher boundary scores (BIoU 54.51 vs. 52.82, BF1 0.6074 vs. 0.5889). 
We conjecture that LCA outperforms CA because CA forces the model to suppress irrelevant distant tokens that diffuse attention away from geometrically relevant context—as in Deformable DETR \cite{zhu2020deformable}, where sparse local attention matched or exceeded its global counterpart. Additionally, since DINO tokens already encode scene-level semantics globally, a few localized tokens may suffice to retrieve the relevant information.
Moreover, We note that when processing 100\% of coordinates our model achieves the best performance, yet with large computational overhead.
By observing the last 4 rows, we can analyze the benefits of using the proposed SEFI strategy. First, by introducing coarse Semantic Edges, $\mathcal{B}_{sem}$, we can achieve a significant decrease in total MACs (3rd vs 5th rows: from 26990G to 7360G). By employing also the image edges, $\mathcal{B}_{img}$ to obtain thin semantic edges we can decrease MACs even more (5th vs last row: 7360G to 6430G), with marginal decrease in performance.
In summary, the semantic-edge inference protocol reduces the number of used coordinates from 100\% to 14\% while preserving precision, substantially improving efficiency at negligible cost to performance.
Overall, these results prove that our method can achieve remarkable performance on very-high-resolution segmentation while remaining efficient by focusing computation where it is most needed.
\begin{table}[t]
\centering
\begin{subtable}[t]{.57\linewidth}\vspace{0pt}
  \centering
  \resizebox{\linewidth}{!}{%
    \begin{tabular}{l|c|c|ccc|cccc}
      \toprule
& \multirow{2}{*}{Encoder} &Encoder & \multicolumn{3}{c|}{MaSS-val (500)} & Model\\
\textbf{Methods} & & Input Res. & mIoU$\uparrow$ & BIoU$\uparrow$ & BF1$\uparrow$  & Params\\
\midrule
MaSSFormer-Lite \cite{Xie2025mass13k} & R18 & $4096 \times 3072$ & 87.11 & 45.35 & .5137 & 15.07M\\
MaSSFormer \cite{Xie2025mass13k} & R50 &$4096 \times 3072$ &  88.97 & 48.97 & .5639  & 37.42M\\
\midrule
DPT \cite{ranftl2021dpt} &DINOv2\texttt{-}S & $2044 \times 2044$ & 90.22 & 45.01  & .5198 &  46.55M\\
DPT \cite{ranftl2021dpt}&DINOv3\texttt{-}S+  & $2048 \times 2048$ & \second{91.88} & 49.30  & .5660 &  53.19M\\
\midrule
\algoname{} & DeiT\texttt{-}S& $2048 \times 2048$ & 90.84 & 50.28 & .5633 & 23.29M\\
\algoname{} & DINOv2\texttt{-}S& $2044 \times 2044$&  91.24 & \second{50.78} &\second{.5685} &  23.29M\\
\algoname{} & DINOv3\texttt{-}S+& $2048 \times 2048$ &\best{\textbf{92.53}} & \best{\textbf{53.24}} & \best{\textbf{.5965}}  &  29.92M\\
\bottomrule
    \end{tabular}%
  }
    \subcaption{\textbf{Decoder Analysis}
}
  \label{tab:back_abl}
\end{subtable}
\begin{subtable}[t]{0.41\linewidth}\vspace{0pt}
  \centering
  \resizebox{\linewidth}{!}{%
    \begin{tabular}{cc|ccc|ccc|c}
\toprule
 & & \multicolumn{3}{c}{\textit{Coarse}} & \multicolumn{3}{|c|}{\textit{Final}} & \\
Stride & Dilation & \multicolumn{3}{c}{MaSS-val (500)}  & \multicolumn{3}{|c|}{MaSS-val (500)} & Coord \\
 $s$ & $r$ & mIoU$\uparrow$ & BIoU$\uparrow$ & BF1$\uparrow$  & mIoU$\uparrow$ & BIoU$\uparrow$ & BF1$\uparrow$  & Usage \%  \\
\midrule
\textbf{8}  &\textbf{5} &\second{90.08}  & \second{27.42} & \second{.2801} &92.53 & 53.24 & .5965 & 14 \\

\midrule
4  &5 & \best{\textbf{91.45}}  & \best{\textbf{42.02}}  & \best{\textbf{.4568}}  &\best{\textbf{92.60}}  &\best{\textbf{54.21}} & \best{\textbf{.6064}} & 17  \\

16  &5 & 88.17 &16.30 & .2065  & 92.20 & 49.64 & .5537 & 15   \\

32 &5 &85.11 & 10.75 & .1669 & 90.96 & 44.78 & .5061 & 18  \\
\midrule
8 &1 &\second{90.08}  & \second{27.42} & \second{.2801} &92.07 & 48.16 & .5433  & 8   \\

8 &3 &\second{90.08}  & \second{27.42} & \second{.2801}  &92.47 & 52.22 & .5822  & 12   \\

8  &7 &\second{90.08}  & \second{27.42} & \second{.2801} & \second{92.56} & \second{53.53} & \second{.5996} & 16  \\
\midrule
32 &7 &85.11 & 10.75 & .1669 &91.47 &46.17 & .5278 & 21\\
32 &15 &85.11 & 10.75 & .1669 &92.30 &51.17 &.5806 & 27\\
\bottomrule
    \end{tabular}%
  }
    \subcaption{\textbf{SEFI Analysis.}
}
      \label{tab:ablation_2}

\end{subtable}
\caption{\textbf{\algoname{} Decoder Analysis (a) on MaSS13K \cite{Xie2025mass13k} and SEFI analysis (b).} “$\uparrow$”  means higher is better. \colorbox{sota}{\strut \textbf{Best}}, \colorbox{secondsota}{\strut \underline{Second-Best}}}
\label{tab:mass13k_analysis}
\end{table}
\paragraph{\textbf{\algoname{} Decoder Contribution Analysis.}}\label{sec:abl_back} 
One could argue that \algoname{}'s performance gains stem primarily from its foundational backbone rather than its architectural design. To isolate the impact of our coordinate-based decoder, we attempted to integrate DINOv3/v2 backbones into competing methods for a fair comparison. However, architectures such as MaSSFormer are specifically tailored for Very-High-Resolution (VHR) data and are not natively compatible with standard foundational ViT encoders. Furthermore, simply introducing a DINO backbone is infeasible in the VHR regime due to prohibitive memory and computational overhead. In contrast, \algoname{} seamlessly integrates any ViT backbone because its coordinate decoder is fully decoupled from the encoder, making the use of foundation models both natural and efficient.
A standard approach for employing foundational encoders in segmentation is to pair them with general dense-prediction decoders, such as DPT \cite{ranftl2021dpt}. Accordingly, we trained DINOv3-S+ \cite{simeoni2025dinov3} and DINOv2-S \cite{oquab2024dinov2} (with registers \cite{darcet2024vitregisters}) using a DPT decoder on MaSS13K, with results reported in \cref{tab:back_abl} (rows 3 and 4). In these experiments, the input resolution is similar to \algoname{} ($\sim2048 \times 2048$); however, we note that larger inputs would be infeasible due to DPT's memory constraints. When comparing these DPT-based results to \algoname{} using the same backbones (last two rows of \cref{tab:back_abl}), \algoname{} achieves significantly higher BIoU and BF1 scores. Remarkably, it does so while requiring nearly half the parameters. This result underscores the importance of our coordinate-based formulation, regardless of the backbone employed.
Finally, we investigate the versatility of the proposed \algoname{} decoder across other pre-trained backbones.
We evaluate our coordinate-based decoder when paired with several encoders: DeiT-S \cite{Touvron2022deits3}, DINOv2-S with registers, and DINOv3-S+ (last three rows of \cref{tab:back_abl}). Our method performs best with strong foundation models, achieving state-of-the-art results with both the DINOv2-S and DINOv3-S+ encoders. Notably, it still surpasses prior methods on most metrics
even with weaker backbones pretrained on ImageNet-1k, such as DeiT-S.
Overall, these results show that our coordinate-based design is both key to performance and broadly applicable.
\paragraph{\textbf{SEFI Analysis.}} In \Cref{tab:ablation_2} we report experimental results that investigates the gain provided by the \emph{Final} inference relatively to the initial \emph{Coarse} segmentation and the impact of changing hyperparameters (stride $s$ and dilation radius $r$) in SEFI.
In general, increasing the coarse stride $s$ reduces coordinate density and degrades boundary alignment. For instance, increasing the stride from $8$ to $32$ substantially degrades the coarse prediction, reducing BIoU to $39\%$ of the baseline (10.75 vs.\ 27.42). This degradation propagates to the final output, whose BIoU drops to $84\%$ of the baseline (44.78 vs.\ 53.24).  These results show that the quality of the coarse segmentation impacts final performance. However, our coordinate-based approach allows increasing the dilation to compensate for less accurate coarse maps. Indeed, with stride $32$, increasing the dilation to $15$ recovers performance, with results slightly below our default setting, still achieving state-of-the-art.
Finally, decreasing the stride to $s=4$, with respect to our default setting ($s=8$), yields further improvements at the cost of processing more coordinates.
\section{Final Discussion}
\label{sec:conclusion}
\paragraph{\textbf{Limitations.}}
We note that \algoname{} can be highly efficient when the hardware provides adequate parallel compute resources (e.g., GPUs with sufficient CUDA cores). On the other hand, when we cannot parallelize computation our method requires to sequentially process coordinates, leading to high inference time. Nevertheless, our method is currently the only one capable of achieving high-quality very-high-resolution segmentations in memory-constraint scenarios.
\paragraph{\textbf{Conclusions.}}
We presented \algoname{}, a coordinate-based framework for very-high-resolution semantic segmentation that combines pixel-level detail with strong global context via a Coordinate Decoder and a Localized Cross-Attention mechanism. This formulation enables precise, flexible, and computationally controllable inference, allowing segmentation at arbitrary resolutions, within regions of interest, or along thin semantic boundaries.
Future extensions include applying our coordinate-based design to instance and panoptic segmentation, and exploring 
dense prediction tasks such as 
depth estimation or image inpainting, where inference can be restricted to task-relevant pixels. We believe these directions highlight the broader potential of coordinate-driven designs for very-high-resolution vision tasks and hope it will inspire further research.
\section*{Acknowledgments}
We acknowledge the CINECA award under the ISCRA initiative, for the
availability of high-performance computing resources and support.

{
    \small
    \bibliographystyle{ieeenat_fullname}
    \bibliography{main}
}
\clearpage
\appendix
\setcounter{section}{0}
\setcounter{table}{0}
\setcounter{figure}{0}
\setcounter{equation}{0}

\renewcommand{\thesection}{A\arabic{section}}
\renewcommand{\thetable}{A\arabic{table}}
\renewcommand{\thefigure}{A\arabic{figure}}
\renewcommand{\theequation}{A\arabic{equation}}
\maketitlesupplementary
\setcounter{page}{1}
\noindent In this supplementary material we provide additional analyses and details of \algoname{}. In particular, the document is organized as follows: 
\begin{itemize}
    \item \textbf{\cref{sec:add_ablation}}. We report additional ablation experiments. Specifically, we analyze how varying the encoder input resolution affects performance and computational cost, we measure the performance gains obtained with a higher-capacity backbone, and we isolate the contribution of each query component.
    \item \textbf{\cref{sec:sup_ade}}. We analyze the annotation quality of a legacy semantic segmentation benchmark (ADE20K \cite{zhou2017ade20k}) compared to MaSS13K \cite{Xie2025mass13k}, and show how \algoname{} performs when annotations are less precise.
    \item \textbf{\cref{sec:sup_implementation}}. To ensure reproducibility, we present a comprehensive overview of the implementation details, including training protocol, optimization hyperparameters, and details on the MACs calculations.
    \item \textbf{\cref{sec:additonal_inf}}. We provide additional details on the inference protocols enabled by \algoname{}.
    \item \textbf{\cref{sec:sup_qual}}. We provide qualitative results on MaSS13K, DIS5K \cite{qin2022dis5k} and \datakidney{} \cite{deng2025kpis}. 
    \item \textbf{\cref{sec:sup_failure}}.
    We present qualitative failure cases of our method on MaSS13K\cite{Xie2025mass13k} and DIS5K\cite{qin2022dis5k}.
\end{itemize}

\section{Additional ablation experiments}
\label{sec:add_ablation}
\subsection{Ablation on the Encoder Input}
\label{sec:sup_input}
The ablation in \cref{tab:supp} examines how the  input resolution of the encoder, $H^\downarrow \times W^\downarrow$, affects accuracy and computational cost on MaSS13K.
In the first two rows, we study the effect of increasing the input resolution of DINOv3 in the baseline configuration that uses only the Segmentation Head (SH) on top of the encoder. Specifically, we raise the input resolution to the maximum allowed by our hardware (i.e., $3072 \times 2304$) while keeping all training settings unchanged. Increasing the resolution from $2048 \times 2048$ to $3072 \times 2304$ yields only modest gains -- especially on boundary metrics (BIoU: $28.72 \rightarrow 32.48$, BF1: $0.3507 \rightarrow 0.3948$) -- while almost tripling the total cost (2960G $\rightarrow$ 7850G MACs). This increase comes almost entirely from the encoder (2950G $\rightarrow$ 7840G MACs), as the SH is extremely lightweight.
The last rows report results for \algoname{} using encoder inputs of $1024 \times 1024$ and $2048 \times 2048$. Unlike DINOv3 + SH, \algoname{} already delivers strong boundary performance when the encoder processes only $1024 \times 1024$ images (91.16 mIoU, 48.83 BIoU, 0.5477 BF1) with a total cost of 3720G MACs. Increasing the encoder resolution to $2048 \times 2048$ leads to clear improvements across all metrics -- particularly on boundaries (mIoU $+1.37$, BIoU $+4.41$, BF1 $+0.0488$) -- while increasing the total cost from 3720G to 6430G MACs (about $1.7\times$).
Crucially, this additional cost is almost entirely absorbed by the encoder (270G $\rightarrow$ 2950G), whereas the decoder remains roughly constant ($\sim $3.4T MACs) thanks to the Coordinate Decoder computational cost being decoupled from the encoder's input size.
\begin{table}[t]
\centering

\resizebox{\linewidth}{!}{
\begin{tabular}{l|c|ccc|cccc}
\toprule
& Encoder & \multicolumn{3}{c|}{MaSS-val (500)} & Model & Total & Encoder & Decoder  \\
\textbf{Methods} & Input Res. & mIoU$\uparrow$ & BIoU$\uparrow$ & BF1$\uparrow$  & Params & MACs & MACs & MACs  \\
\midrule
DINOv3 \cite{simeoni2025dinov3} + SH & $2048 \times 2048$  & 90.09 & 28.72 &  .3507 & 28.70M & 2960 G & 2950 G &  10 G \\
DINOv3 \cite{simeoni2025dinov3} + SH & $3072 \times 2304$  & 90.36 & 32.48  & .3948 & 28.70M & 7850 G & 7840 G &  10 G \\
\midrule
\algoname{} & $1024 \times 1024$  & \second{91.16} & \second{48.83} & \second{.5477} &29.92M & 3720 G & 270 G & 3450 G \\
\algoname{} & $2048 \times 2048$  & \best{\textbf{92.53}} & \best{\textbf{53.24}} & \best{\textbf{.5965}} &  29.92M& 6430 G & 2950 G & 3480 G  \\

\bottomrule
\end{tabular}}
\caption{\textbf{Analysis on the Encoder Input Size on MaSS13K\cite{Xie2025mass13k}.} “$\uparrow$”  means higher is better.\colorbox{sota}{\strut \textbf{Best}}, \colorbox{secondsota}{\strut \underline{Second-Best}}}

\label{tab:supp}
\end{table}






\begin{table}[t]
\centering

\renewcommand{\arraystretch}{1.3}
\resizebox{0.9\linewidth}{!}{
\begin{tabular}{lc|ccc|ccc|c}
\toprule
 & 
 & \multicolumn{3}{c|}{MaSS-val (500)}
 & \multicolumn{3}{c|}{MaSS-test (1,500)}
 & \multicolumn{1}{c}{Stat.} \\
 \textbf{Methods}&Backbone & mIoU$\uparrow$ & BIoU$\uparrow$ & BF1$\uparrow$
   & mIoU$\uparrow$ & BIoU$\uparrow$ & BF1$\uparrow$
   & Param. \\
\midrule
MaSSFormer-Lite \cite{Xie2025mass13k} & R18 & 87.11 & 45.35 & .5137 & 86.13 & 43.28 & .5086 & 15.07M\\
MaSSFormer \cite{Xie2025mass13k} & R50 & 88.97 & 48.97 & .5639 & 88.21 & 48.39 & .5593 & 37.42M \\
\midrule
\algoname{}        & DINOv3\texttt{-}S+ & \second{92.53} & \second{53.24} & \second{.5965} & \second{92.30} & \second{52.40} & \second{.5916} & 29.92M\\
\algoname{}\texttt{-}B   & DINOv3\texttt{-}B  & \best{\textbf{92.59}} & \best{\textbf{54.65}} & \best{\textbf{.6163}} & \best{\textbf{92.68}}& \best{\textbf{54.10}} & \best{\textbf{.6109}} & 90.38M\\
\cline{1-9}
\end{tabular}}
\caption{\textbf{Analysis on the Encoder Capacity on MaSS13K \cite{Xie2025mass13k}.} “$\uparrow$” means higher is better.\colorbox{sota}{\strut \textbf{Best}}, \colorbox{secondsota}{\strut \underline{Second-Best}}}
\label{tab:supp_scala}

\end{table}
\begin{figure*}[t]
    \centering

    \begin{subfigure}{0.46\linewidth}
        \centering
        \includegraphics[width=0.31\linewidth]{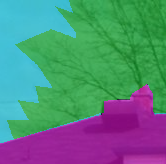}
        \includegraphics[width=0.31\linewidth]{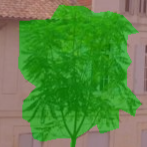}
        \includegraphics[width=0.31\linewidth]{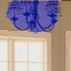}
        \caption{Groundtruth ADE20K}
        \label{fig:dataset_b}
    \end{subfigure}
    \hspace{0.01
    \linewidth}
    \begin{subfigure}{0.46\linewidth}
        \centering
        \includegraphics[width=0.31\linewidth]{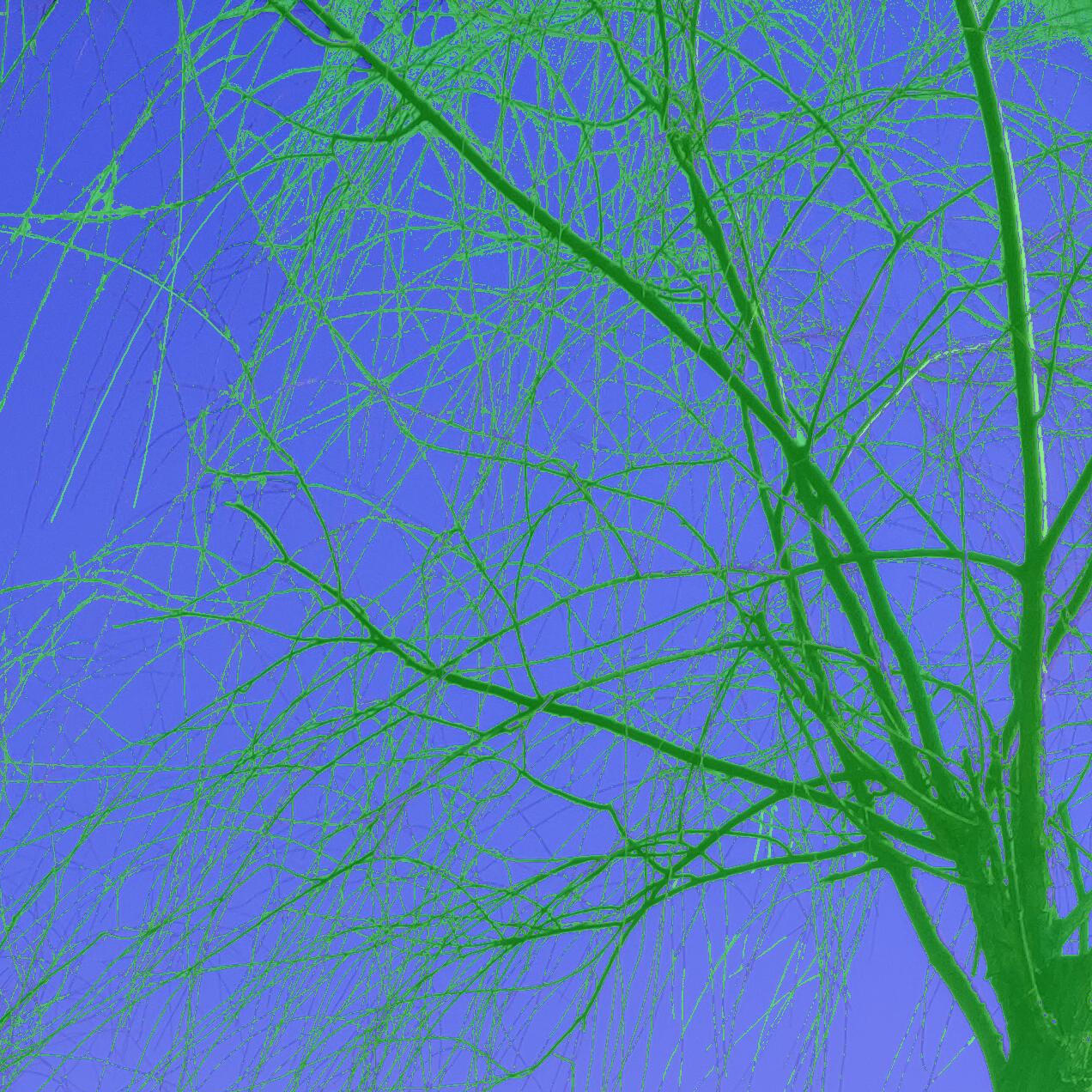}\
        \includegraphics[width=0.31\linewidth]{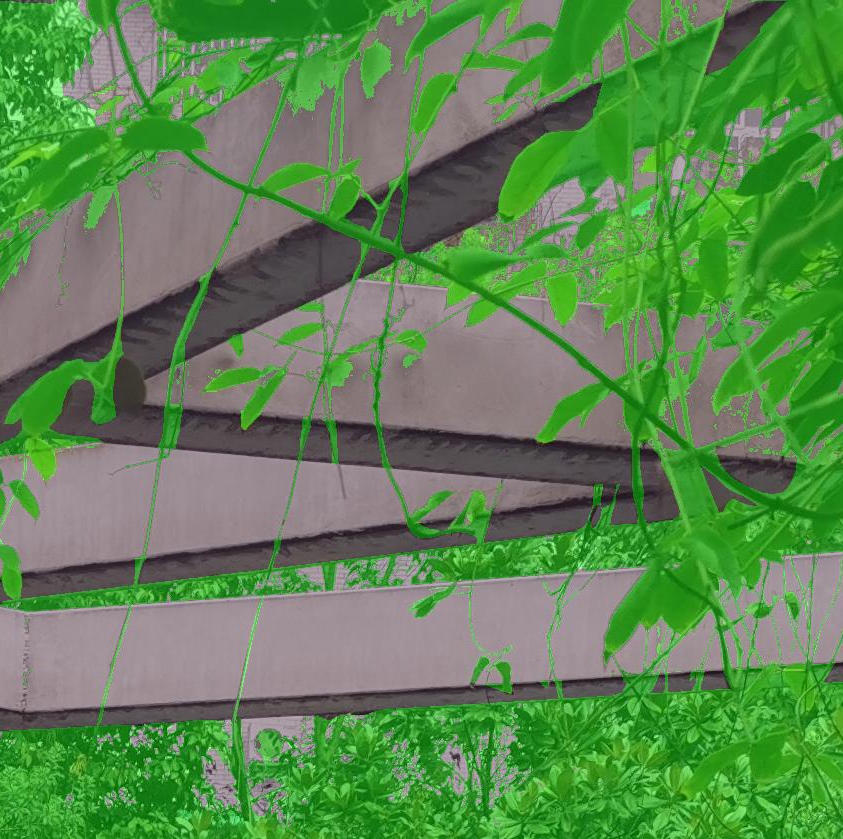}
        \includegraphics[width=0.31\linewidth]{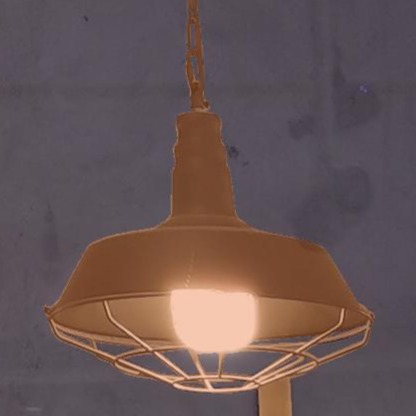}
        \caption{Groundtruth MaSS13K}
        \label{fig:dataset_a}
    \end{subfigure}
    \caption{\textbf{Annotation sharpness comparison:} ADE20K \cite{zhou2017ade20k} (a) vs MaSS13K \cite{Xie2025mass13k} (b) }
    \label{fig:comparison}
\end{figure*}
\subsection{Ablation on the Encoder Capacity}
\label{sec:scalability}
We investigate how the capacity of the encoder affects the quality of our predictions. Starting from our default configuration, which builds on DINOv3\texttt{-}S+, we replace the backbone with the deeper DINOv3\texttt{-}B and train the model under the same protocol, denoting this variant \algoname{}\texttt{-}B. \cref{tab:supp_scala} reports the results on both the MaSS-val and MaSS-test splits of MaSS13K~\cite{Xie2025mass13k}.
The deeper backbone yields consistent improvements across all metrics. On MaSS-test, \algoname{}\texttt{-}B improves the mIoU by $0.38$ points over our default model, while the boundary-aware metrics benefit substantially more, with BIoU rising by $1.70$ points and BF1 by $1.9$ points. The same trend holds on MaSS-val. This asymmetry is informative: the region-level mIoU is already close to saturation in our default configuration, so the additional encoder capacity translates primarily into sharper, more accurate object boundaries rather than into coarse region accuracy. This is consistent with the design of our method, whose contribution is concentrated precisely at the boundaries. It is also worth noting that even our default DINOv3\texttt{-}S+ model already surpasses MaSSFormer~\cite{Xie2025mass13k} by a large margin on every metric while using fewer parameters ($29.92$M against $37.42$M), and \algoname{}\texttt{-}B further widens this gap at the cost of a roughly threefold increase in parameters ($90.38$M). The deeper variant therefore, offers a favorable accuracy–complexity trade-off when boundary fidelity is the priority, whereas the default configuration remains the most efficient choice for general use.
\begin{table}[t]
\centering
\resizebox{0.6\linewidth}{!}{
\renewcommand{\arraystretch}{1.2}
\begin{tabular}{lcc|ccc}
\toprule
 & & & \multicolumn{3}{c}{MaSS-val (500)} \\
\textbf{Methods} & $f_c$ & $f_p$ & mIoU $\uparrow$ & BIoU $\uparrow$ & BF1 $\uparrow$ \\
\midrule
\algoname{} & \cmark & \xmark & 91.55 & 42.75 & 49.39 \\
\algoname{} & \xmark & \cmark & \second{92.15} & \second{52.97} & \second{59.25} \\
\algoname{} & \cmark & \cmark & \best{\textbf{92.53}} & \best{\textbf{53.24}} & \best{\textbf{59.65}} \\
\bottomrule
\end{tabular}}
\caption{\textbf{Query components ablation on MaSS13K \cite{Xie2025mass13k}.} Contribution of the coordinate embedding $f_c$ and the local patch embedding $f_p$. “$\uparrow$” means higher is better.}
\label{tab:query_ablation}
\end{table}
\subsection{Ablation on the Query Components}
\label{sec:query_ablation}
The query fed to our decoder combines a local patch embedding $f_p$ with a coordinate embedding $f_c$ through $\phi([f_p, f_c])$. To isolate the contribution of each component, we evaluate three variants: the coordinate embedding alone ($f_c$), the patch embedding alone ($f_p$), and the full query ($\phi([f_p, f_c])$). All variants share the same backbone and training protocol, and we report mIoU together with the boundary-sensitive BIoU and BF1 metrics on MaSS13K \cite{Xie2025mass13k}. Results are shown in \cref{tab:query_ablation}. The local patch embedding is the dominant source of information: using $f_p$ alone reaches 52.97 BIoU and 59.25 BF1, far above the 42.75 BIoU and 49.39 BF1 obtained from the coordinate embedding alone. The gap is most pronounced on the boundary metrics, confirming that the high-resolution local appearance captured by $f_p$ is what drives boundary accuracy, whereas the coordinate embedding cannot localize boundaries on its own. The two components are nonetheless complementary: the full query $\phi([f_p, f_c])$ achieves the best result across all three metrics (92.53 mIoU, 53.24 BIoU, 59.65 BF1), showing that the coordinate embedding contributes positional information that refines the prediction beyond what local appearance alone provides.
\section{Evaluation on ADE20K}
\label{sec:sup_ade}
We choose not to focus on standard semantic segmentation benchmarks such as ADE20K, since they do not provide high-resolution images paired with accurate, fine-grained annotations. This is apparent from \cref{fig:comparison}, which visually compares the annotations of ADE20K and MaSS13K. For similar objects, the annotation sharpness differs markedly between the two: ADE20K prioritizes covering a large number of classes, whereas MaSS13K emphasizes precise, accurate delineation. Our method is designed to address the challenges of MaSS13K (very-high-resolution images and boundary accuracy) rather than those of ADE20K (mainly large class diversity and scale variation).\\
Nonetheless, we report experiments on ADE20K in \cref{tab:result_ade}, evaluated with the standard mIoU metric. As shown in the table, \algoname{} reaches 49.7 mIoU with only 29.9M parameters, outperforming all methods built on small-scale (S) backbones---including UPerNet with DeiT-Adapter-S, which uses nearly twice as many parameters (58M)---and trailing only the substantially heavier models equipped with Swin-B and Swin-L backbones. We also evaluate \algoname{} with a DINOv3\texttt{-}L backbone (last row), which achieves 57.0 mIoU, the highest among all compared methods. These results confirm that, although ADE20K cannot fully reflect the boundary precision \algoname{} is designed for, our method remains competitive on a benchmark outside its intended setting.

\begin{table}[t]\small
\centering

\resizebox{0.6\linewidth}{!}{
\begin{tabular}{l|c|c|c|cc}
\toprule
\textbf{Methods} &Backbone & Param. & mIoU $\uparrow$ \\
\midrule
Semantic FPN ~\cite{kirillov2019panoptic}& PVT-Small ~\cite{wang2021pyramid}  & 28.2M & 41.9 \\
 Semantic FPN ~\cite{kirillov2019panoptic} & PVTv2-B2 ~\cite{wang2021pvtv2}  & 29.1M & 45.2 \\
Semantic FPN ~\cite{kirillov2019panoptic} & Swin-T ~\cite{liu2021swin}   & 31.9M & 41.5 \\
Semantic FPN ~\cite{kirillov2019panoptic} & Twins-SVT-S ~\cite{chu2021twins}  & 28.3M & 43.2 \\
		
Semantic FPN ~\cite{kirillov2019panoptic} & ViT-S~\cite{li2021benchmarking}  & 27.8M & 44.6  \\
UperNet ~\cite{xiao2018unified}  & DeiT-Adapter-S ~\cite{chen2023vitadapter} & 58M  & 46.2 \\
MaskFormer ~\cite{cheng2021maskformer}  & Swin-B ~\cite{liu2021swin}      & 102M & 52.7\\
Mask2Former ~\cite{cheng2022mask2former} & Swin-B ~\cite{liu2021swin}        & 107M & 53.9\\
MaskFormer ~\cite{cheng2021maskformer}  & Swin-L ~\cite{liu2021swin}        & 212M & 54.1\\
Mask2Former ~\cite{cheng2022mask2former} & Swin-L ~\cite{liu2021swin}       & 216M & \second{56.1}\\
\midrule
DINOv3 ~\cite{simeoni2025dinov3} + SH &DINOv3\texttt{-}S+& 28.7M &  47.7 \\
\midrule
\algoname{}& DINOv3\texttt{-}S+& 29.9M & 49.7 \\
\algoname{}\texttt{-}L & DINOv3\texttt{-}L  & 309M  & \best{\textbf{57.0}}\\


		\bottomrule
	\end{tabular}

}
\caption{\textbf{Semantic segmentation results on ADE20K \cite{zhou2017ade20k} val.} “$\uparrow$” means higher is better.\colorbox{sota}{\strut \textbf{Best}}, \colorbox{secondsota}{\strut \underline{Second-Best}}}
\label{tab:result_ade}
\end{table}
\begin{table*}[t]

    \centering
    \label{tab:coordformer_architecture}
    \resizebox{\linewidth}{!}{%
    \begin{tabular}{llll}
        \toprule
        Stage & Layer / Operation & Input $\rightarrow$ Output & Description \\
        \midrule
        \multirow{3}{*}{\shortstack[l]{Encoder \\ (e.g., DINOv3-S+)}}
        & Image downsampling
        & $I \in \mathbb{R}^{H \times W \times 3}
           \rightarrow I^{\downarrow} \in \mathbb{R}^{H^{\downarrow} \times W^{\downarrow} \times 3}$
        & Very-high-resolution RGB image is resized
          (e.g., $H^{\downarrow}=W^{\downarrow}=2048$). \\
        & ViT backbone $\mathcal{E}$
        & $I^{\downarrow} \rightarrow \{F_i\}_{i=1}^{4}$
        & Extract from DINOv3-S+ four
          $F_i \in \mathbb{R}^{h \times w \times c_e}$ from transformer blocks. \\
        & Multi-level token concat.
        & $\{F_i\}_{i=1}^{4} \rightarrow F_g$
        & After projecting $F_i$ to $F_i'$ concatenate along channels:
          $F_g = [F_1', F_2', F_3', F_4'] \in \mathbb{R}^{h \times w \times 4c_l}$ \\
        \midrule
        \multirow{4}{*}{\shortstack[l]{Coordinate \\ Decoder}}
        & Patch extraction
        & $(I, (x,y)) \rightarrow P_{x,y}$
        & Extract a local patch
          $P_{x,y}$ of size $p \times p$ ($p = 8$) from $(x,y)$
          in the original image. \\
        & Patch MLP $\Psi$
        & $P_{x,y} \rightarrow f_p$
        & Patch embedder
          $\Psi: \mathbb{R}^{3p^2} \rightarrow \mathbb{R}^{d_p}$ encodes fine details into a patch feature $f_p$. \\
        & frequency transform $\Omega$
        & $(x,y) \rightarrow f_c$
        & Normalize $(x,y)$ to $[0,1]^2$ and map with
          $\Omega: \mathbb{R}^2 \rightarrow \mathbb{R}^{d_c}$ to obtain $f_c$. \\
        & projection $\phi$
        & $[f_p, f_c] \rightarrow f_q$
        & Concatenate $f_p$ and $f_c$ and project
          $\phi: \mathbb{R}^{d_p + d_c} \rightarrow \mathbb{R}^{c_q}$ to form the query $f_q$. \\
        \midrule
        \multirow{3}{*}{\shortstack[l]{Localized \\ Cross-Attention \\ (LCA)}}
        & Coordinate mapping
        & $(x,y) \rightarrow (x^{\downarrow}, y^{\downarrow})$
        & $(x^{\downarrow},y^{\downarrow}) = \left(
          \tfrac{x}{s_W \, p_e},\tfrac{y}{s_H \, p_e}\right)$,
          $s_H,s_W$ are downsampling factors and $p_e$ is the patch size. \\
        & Local token selection
        & $(F_g, (x^{\downarrow}, y^{\downarrow})) \rightarrow F_{\text{local}}$
        & Select local neighborhood of $k=4$ tokens around
          $(x^{\downarrow},y^{\downarrow})$ in $F_g$ to obtain tokens $F_{\text{local}}$. \\
        & Cross-Attention
        & $(f_q, F_{\text{local}}) \rightarrow f'_q$
        & $f_q$ as query and
          $F_{\text{local}}$ as keys/values, refined token
          $f'_q$ with global context. \\
        \midrule
        Segmentation Head
        & MLP head (SH)
        & $f'_q \rightarrow \hat{z}_{x,y}$
        & MLP maps
          $f'_q \in \mathbb{R}^{c_q}$ to class logits
          $\hat{z}_{x,y} \in \mathbb{R}^{K}$ for the coordinate. \\
        \bottomrule
    \end{tabular}%
    }
    \caption{\textbf{COORDFORMER architecture overview.}
    \label{tab:architecture_supp_}
    The table details the subsequent
    coordinate-based decoding pipeline, from input image and coordinates to
    final class logits. Notation follows Sec.~3.1.}
\end{table*}
\begin{algorithm*}[t]
\caption{Semantic-Edge-Focused Inference (SEFI)}
\label{alg:sefi}
\KwIn{
High-res image $I\in\mathbb{R}^{H\times W\times 3}$;
\algoname{} $f_\theta$;
coarse stride $s$;
dilation radius $r$;
Sobel threshold $\tau$ (default $0.1$);
coordinate batch size $B$.
}
\KwOut{Dense label map $\hat{\mathbf{Y}}\in\{1,\dots,K\}^{H\times W}$.}

\BlankLine
\textbf{(1) Coarse querying at stride $s$}\;
$\mathcal{C}\leftarrow\{(x,y)\,:\, x\in\{0,s,2s,\dots\},\, y\in\{0,s,2s,\dots\}\}$\;
$\hat{\mathbf{Z}}_{c} \leftarrow \textsc{QueryLabels}(f_\theta,I,\mathcal{C})$ \tcp*[r]{reshape on coarse grid $\lceil H/s\rceil\times \lceil W/s\rceil$}

$\hat{\mathbf{Z}}_{0} \leftarrow \textsc{Upsample}(\hat{\mathbf{Z}}_{c}, H,W)$\;

\BlankLine
\textbf{(2) Semantic boundary mask from coarse prediction}\;
$\mathcal{B}_{\text{sem}} \leftarrow \textsc{SemanticBoundary}(\hat{\mathbf{Z}}_{c})$\;
$\mathcal{B}_{\text{sem}} \leftarrow \textsc{Dilate}(\mathcal{B}_{\text{sem}}, r)$\;
$\mathcal{B}^{\uparrow}_{\text{sem}} \leftarrow \textsc{Upsample}(\mathcal{B}_{\text{sem}}, H,W)$\;

\BlankLine
\textbf{(3) Image boundary mask (Sobel)}\;
$\mathcal{B}_{\text{img}} \leftarrow \mathbbm{1}\big[\textsc{Sobel}(I)>\tau\big]$\;

\BlankLine
\textbf{(4) Refined semantic boundary coordinates}\;
$\mathcal{B} \leftarrow \mathcal{B}^{\uparrow}_{\text{sem}} \land \mathcal{B}_{\text{img}}$\;
$\mathcal{Q}\leftarrow\{(x,y)\,:\,\mathcal{B}(x,y)=1\}$\;

\BlankLine
\textbf{(5) Boundary refinement by coordinate querying}\;
\ForEach{mini-batch $\mathcal{Q}_b \subseteq \mathcal{Q}$ of size $B$}{
    $\hat{z}_b \leftarrow \textsc{QueryLabels}(f_\theta,I,\mathcal{Q}_b)$\;
    $\hat{\mathbf{Z}}_{0}[\mathcal{Q}_b] \leftarrow \hat{z}_b$ \tcp*[r]{overwrite boundary in the upsampled coarse map}
}
${\hat{\mathbf{Y}}_{0}}\leftarrow \arg\max_k \hat{\textbf{Z}}^k_0$\;
\Return{$\hat{\mathbf{Y}}_{0}$}\;

\BlankLine
\textbf{Subroutine:} $\textsc{QueryLabels}(f_\theta,I,\mathcal{S})$\;
\hspace{1.2em}For each $(x,y)\in\mathcal{S}$, compute logits $\hat{z}_{x,y}=f_\theta(I,(x,y))$ 
\end{algorithm*}    
\begin{figure*}[t]
  \centering
  \begin{subfigure}{\linewidth}
  \begin{subfigure}[t]{0.3\linewidth}
    \centering
    \includegraphics[width=\linewidth]{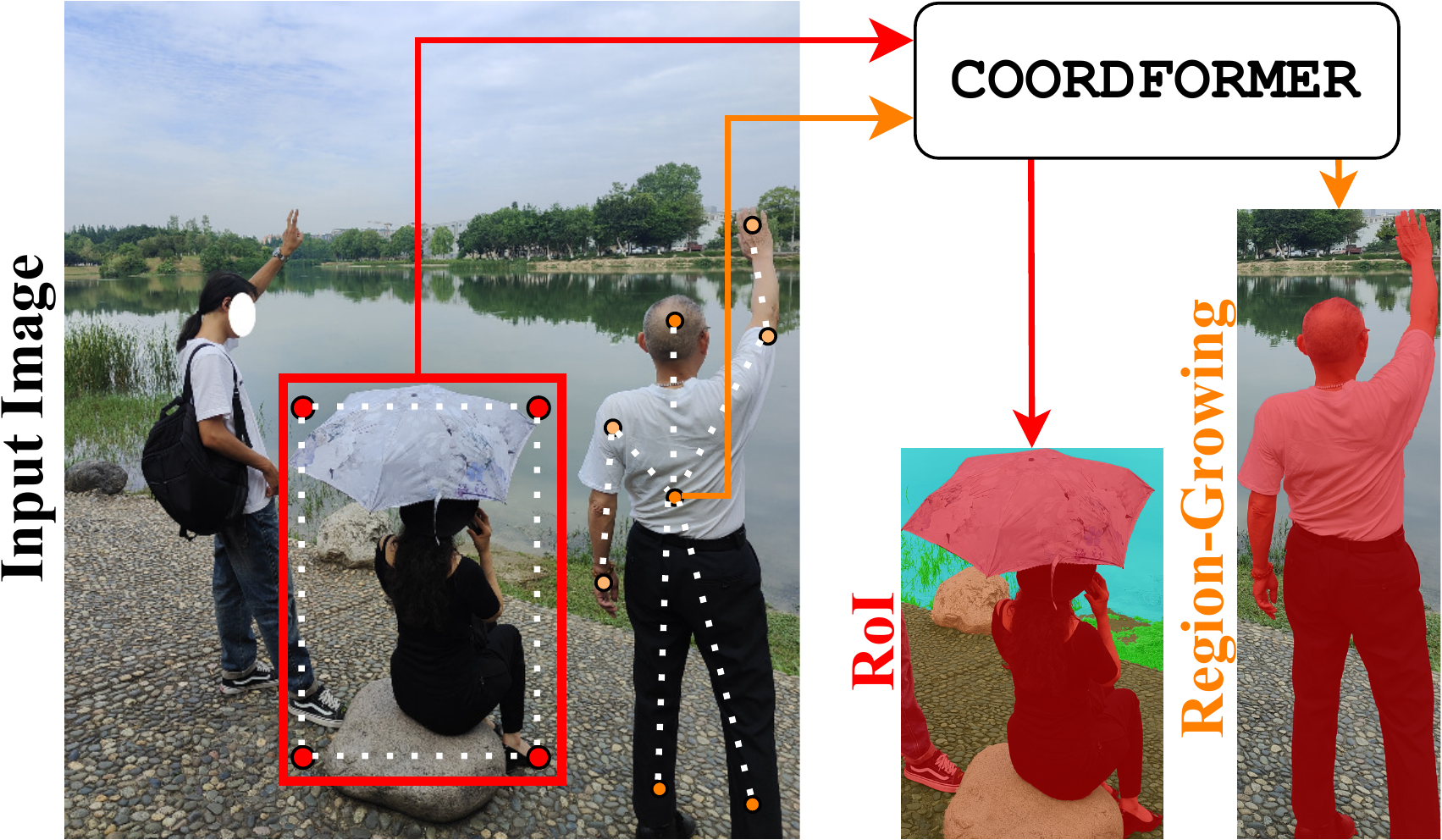}
    \subcaption{\textbf{RoI and Region-Growing Inference}}
    \label{fig:inference_protocols}
  \end{subfigure}
  \hfill
  \begin{subfigure}[t]{0.64\linewidth}
    \centering
    \includegraphics[width=\linewidth]{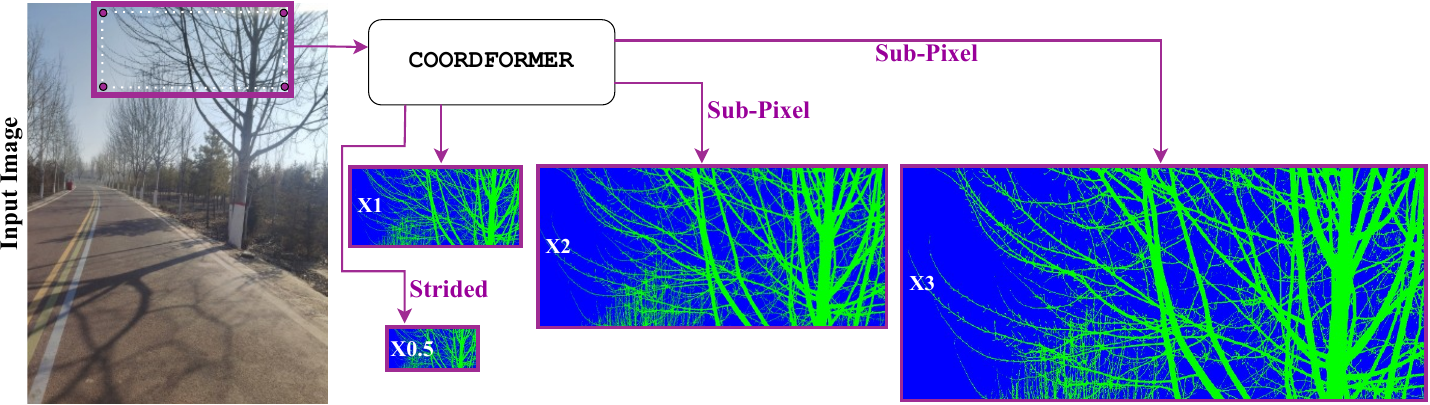}
    \subcaption{\textbf{Sub-Pixel and Strided Inference} }
    \label{fig:super_res}
  \end{subfigure}
\end{subfigure}
  \caption{\textbf{Inference protocols enabled by \algoname{}.} Beyond the Semantic-Edge-Focused Inference described in the main paper, \algoname{} supports several other inference protocols: RoI querying, region growing, sub-pixel, and strided inference.} 
  \label{fig:diff_inf}
\end{figure*}
\section{Additional Implementation Details}
\label{sec:sup_implementation}
\algoname{} is implemented on top of \texttt{semantic-segmentation} library\footnote{\href{https://github.com/sithu31296/semantic-segmentation}{https://github.com/sithu31296/semantic-segmentation}} and PyTorch 2.6. We adopt mixed precision training with bfloat16 autocasting. We train the model using the AdamW optimizer with a batch size of 16. The initial learning rate is set to $1e^{-4}$ with a weight decay of $1e^{-4}$, and we apply a linear decay schedule after a linear warmup over the first ten epochs. For data augmentation, we use random flipping and random photometric distortion. The number of MACs for all variants of \algoname{} is computed using the tools provided in \texttt{PyTorch}. We also adopt an early stopping strategy, terminating training when no improvement in the validation metric is observed for several consecutive epochs. The weights for the Cross-Entropy Loss, Binary Dice Loss and Binary Cross-Entropy Loss, $\lambda_{\mathrm{CE}}$, $\lambda_{\mathrm{D}}$ and $\lambda_{\mathrm{BCE}}$, are set to 2, 5 and 5, respectively. Moreover, in \cref{tab:architecture_supp_}, we report an architecture overview. The encoder tokens $F_i$ are extracted from multiple depths of the transformer encoder $\mathcal{E}$ at regular layer intervals. For instance, DINOv3-S+ has 12 layers, so we take features every three layers (i.e., from four evenly spaced layers).
We highlight that $\Omega$ allows to obtain Fourier-feature embedding \cite{tancik2020fourier} to lift 2D coordinates into a higher-dimensionality. We use 10 frequencies in our implementation.
The projection layers, that process $F_i$ to obtain $F_i'$, reduce the number of channels of the feature map from $c_e=384$ to $c_l = 96$.
The Patch MLP $\Psi$ is implemented as a standard ViT-like patch embedder \cite{dosovitskiy2020vit}, using one convolutional layer with kernel size $8$, stride $8$, and $d_p = 384$.
The block $\phi$, is a sequence of linear layer and ReLU activations which projects the concatenation of patch and coordinate tokens to the query dimension $c_q=384$.
The Localized Cross-Attention (LCA) layer is implemented as a standard cross-attention block between the coordinate-aware queries and the selected local encoder tokens, omitting the feed-forward network typically applied after the attention operation. The threshold used to obtain $\mathcal{B}$ is set to 0.1.
%
Regarding DINOv3 \cite{simeoni2025dinov3}, DINOv2\cite{oquab2024dinov2}, DeiT-S \cite{Touvron2022deits3}, and MaSSFormer \cite{Xie2025mass13k}, we used the official codes and weights available online. We also provide the pseudocode for the SEFI technique in \cref{alg:sefi}.
\section{\textbf{Additional Inference Protocols}}
\label{sec:additonal_inf}
Our method enables several inference protocols (\cref{fig:diff_inf}). While Semantic-Edge-Focused Inference is described in Sec. 3.3 of the main paper, \algoname{} also supports other protocols, some of which substantially reduce unnecessary computation on very-high-resolution images. As shown in \cref{fig:inference_protocols}, our framework can process only the regions or coordinates of interest rather than the full image.
For instance, one can extract an arbitrary Region of Interest (RoI) and segment it at full resolution by querying only its pixel coordinates, yielding detailed local predictions at a fraction of the global cost. The method also supports interactive, category-aware Region Growing: starting from a single user-selected pixel, we iteratively query neighboring coordinates and expand the region until a different semantic category is encountered, producing coherent semantic regions without processing the entire image.
Moreover, \algoname{} allows the output resolution to be chosen (\cref{fig:super_res}) by sampling coordinates at any density, enabling coarse (Strided Inference), medium, or fully detailed masks depending on computational needs. 
Finally, predicting labels from continuous coordinates lets \algoname{} query sub-pixel locations (Sub-Pixel Inference), yielding masks at resolutions beyond the input image---a property that could benefit downstream tasks such as mask-conditioned generation and restoration.
\section{Additional Qualitative Results}
\label{sec:sup_qual}
\begin{figure*}
    \centering
    \includegraphics[width=0.75\linewidth]{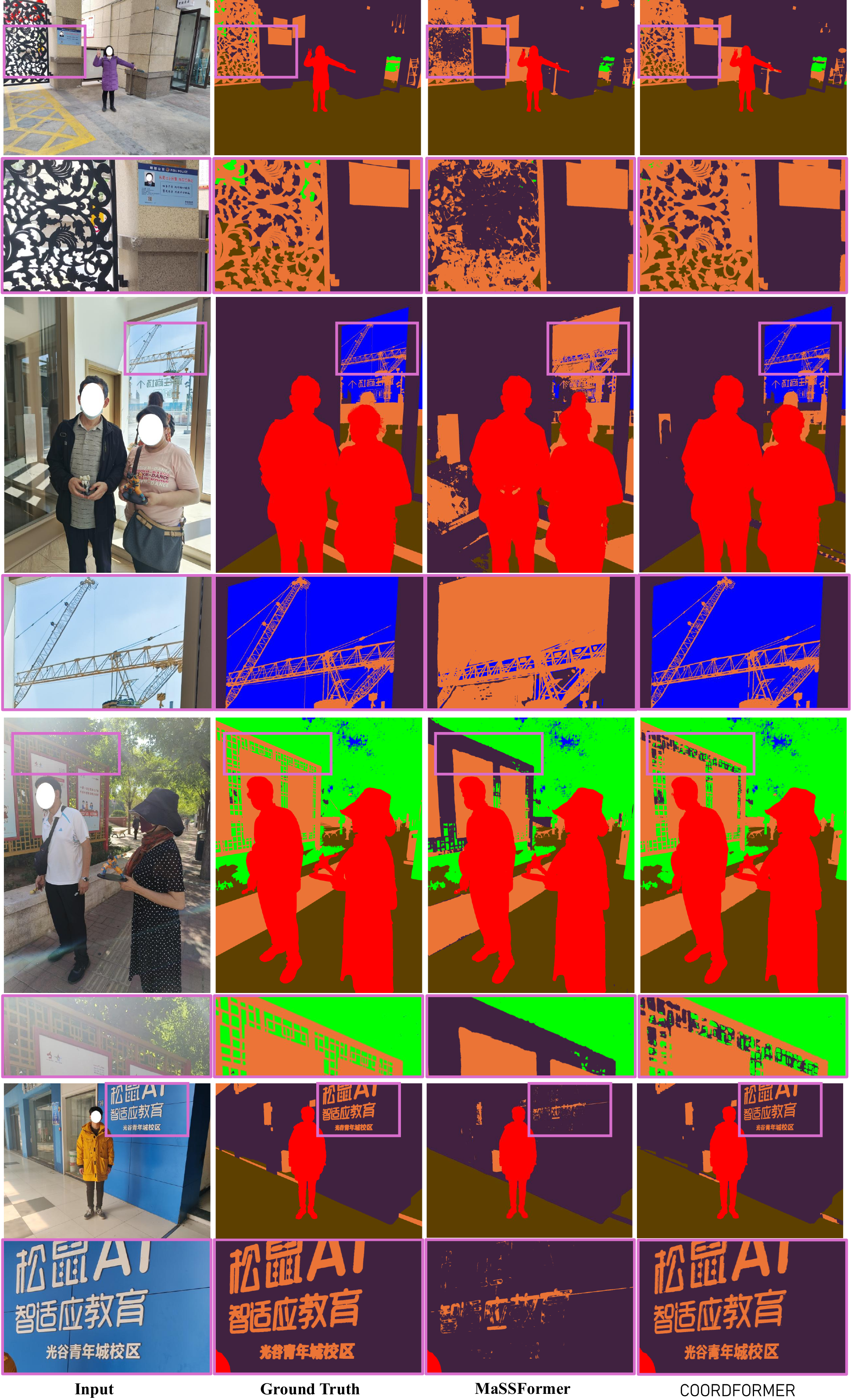}
    \caption{\textbf{Qualitative comparison between \algoname{} and MaSSFormer on MaSS13K \cite{Xie2025mass13k}.} Please zoom in for a clearer view.}
    \label{fig:qualitative_supp_mass}
\end{figure*}

\begin{figure*}
    \centering
    \includegraphics[width=0.86\linewidth]{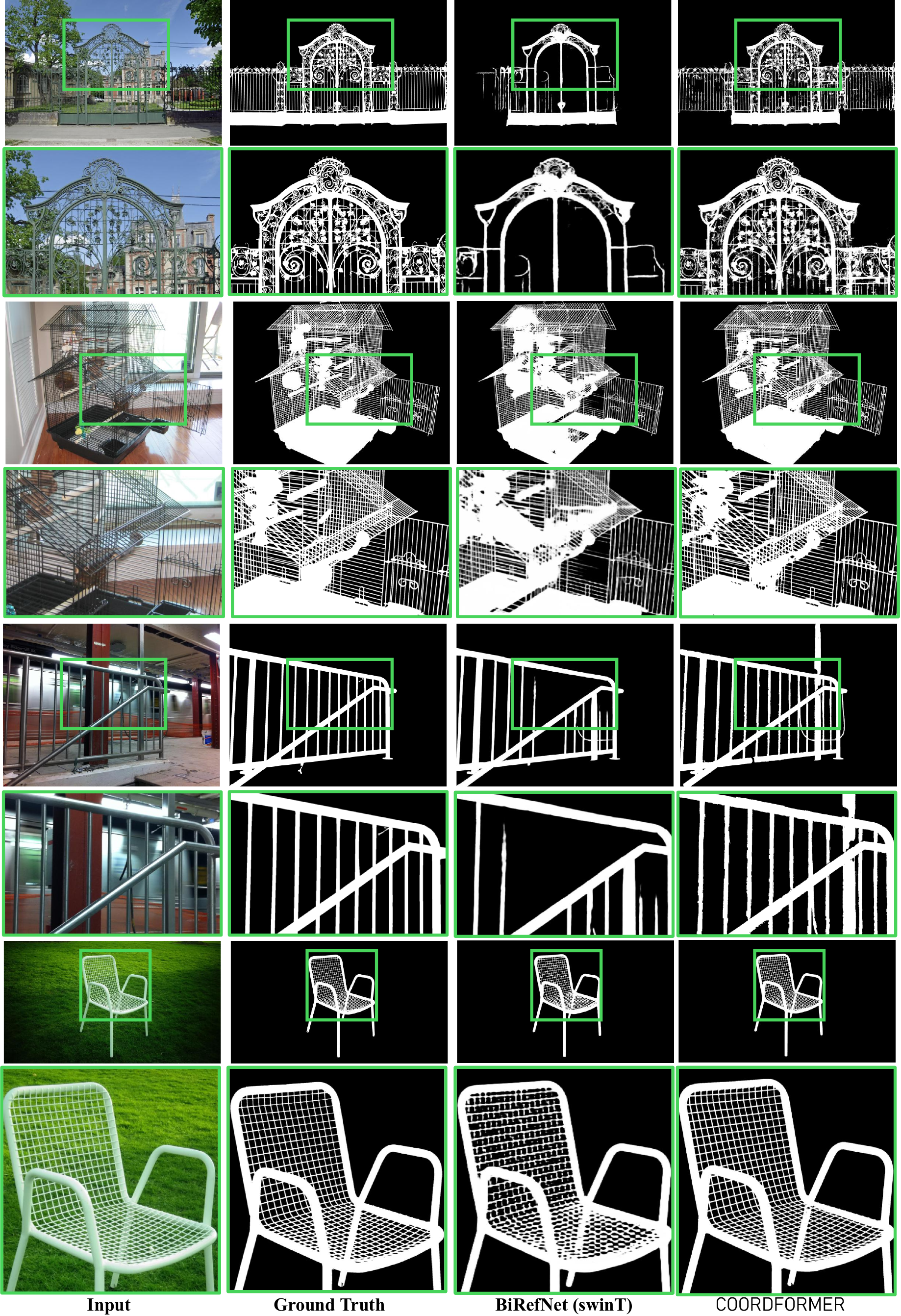}
    \caption{\textbf{Qualitative comparison between \algoname{} and BiRefNet on DIS5K \cite{qin2022dis5k}}. Please zoom in for a clearer view.}
    \label{fig:qualitative_supp_dis}
\end{figure*}

\begin{figure*}
    \centering
    \includegraphics[width=0.58\linewidth]{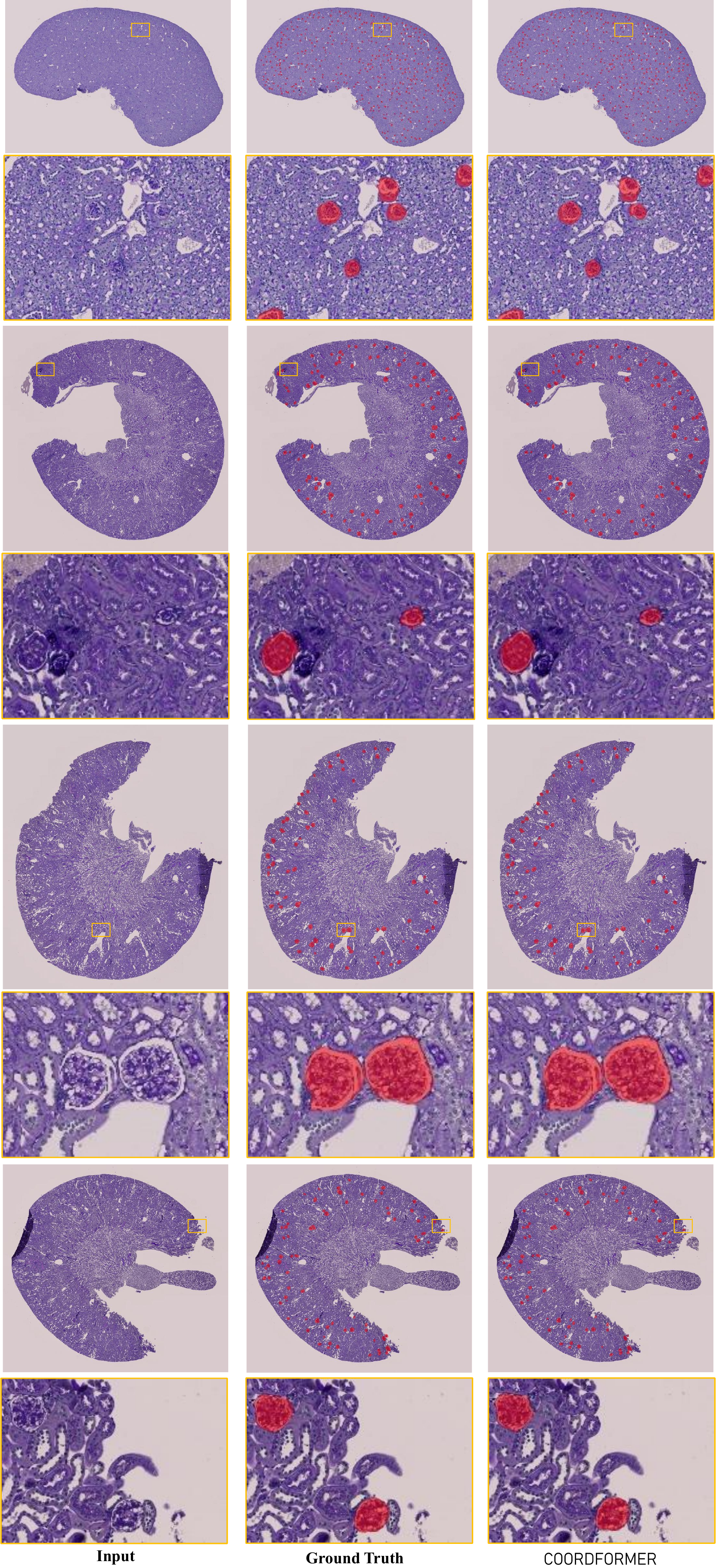}
    \caption{\textbf{Qualitative comparison of \algoname{} and Ground Truth on \datakidney \cite{deng2025kpis}}. Please zoom in for a clearer view.}
    \label{fig:qualitative_supp_kpis}
\end{figure*}

Additional qualitative results for MaSS13K~\cite{Xie2025mass13k}, DIS5K~\cite{qin2022dis5k}, and \datakidney~\cite{deng2025kpis} are shown in \cref{fig:qualitative_supp_mass}, \cref{fig:qualitative_supp_dis}, and \cref{fig:qualitative_supp_kpis}, respectively. 
The images have been downsampled for PDF size constraints. Some of the original resolution predictions are included in the zip file attached with the submission.
\section{Failure Cases}\label{sec:sup_failure}
\begin{figure*}
    \centering
    \includegraphics[width=0.86\linewidth]{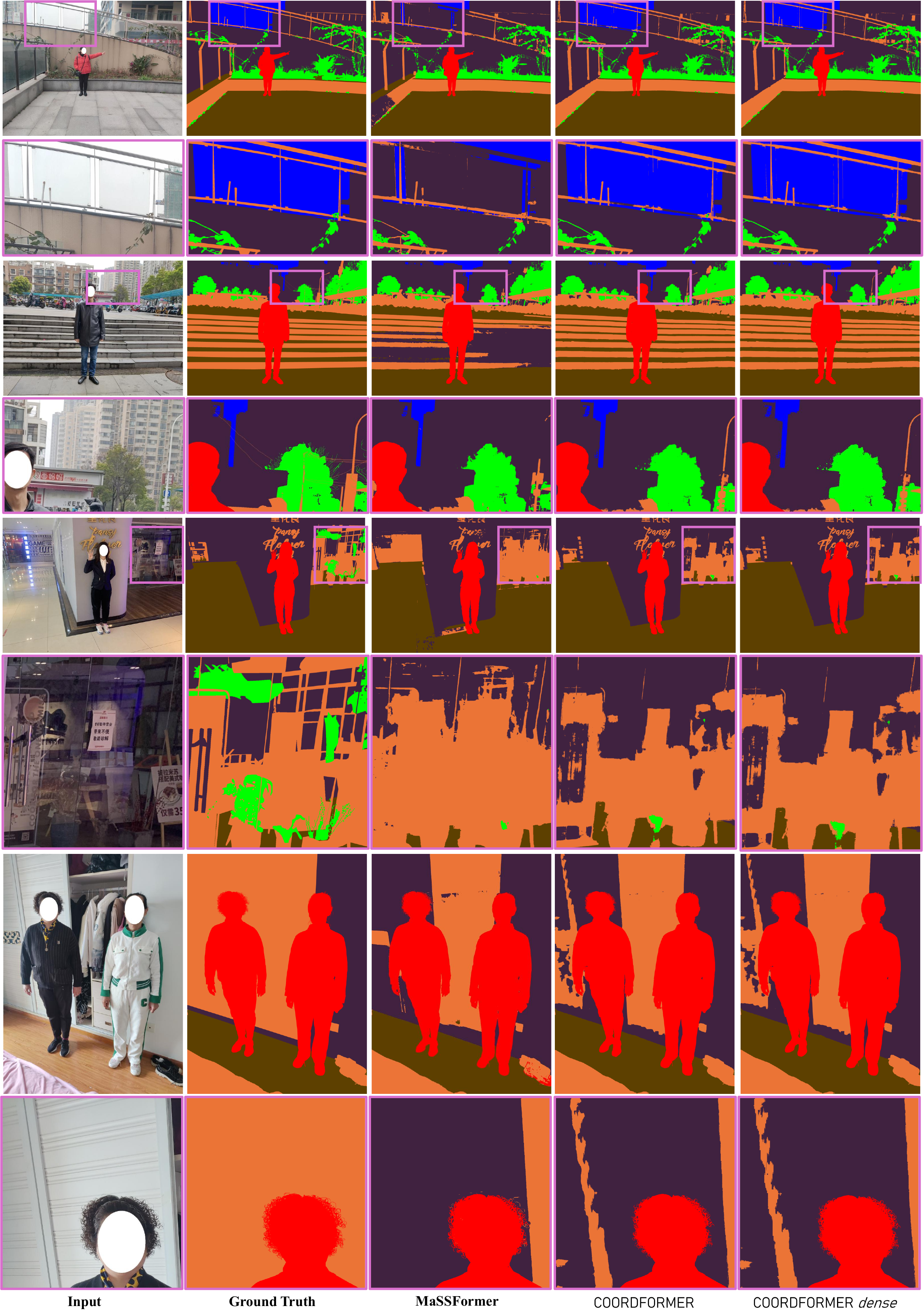}
    \caption{\textbf{Failure cases of \algoname{}. Our method is compared with \algoname{} \textit{dense} and MaSSFormer on MaSS13K \cite{Xie2025mass13k}}. Please zoom in for a clearer view.}
    \label{fig:qualitative_supp_mass_fail}
\end{figure*}

\begin{figure*}
    \centering
    \includegraphics[width=0.8\linewidth]{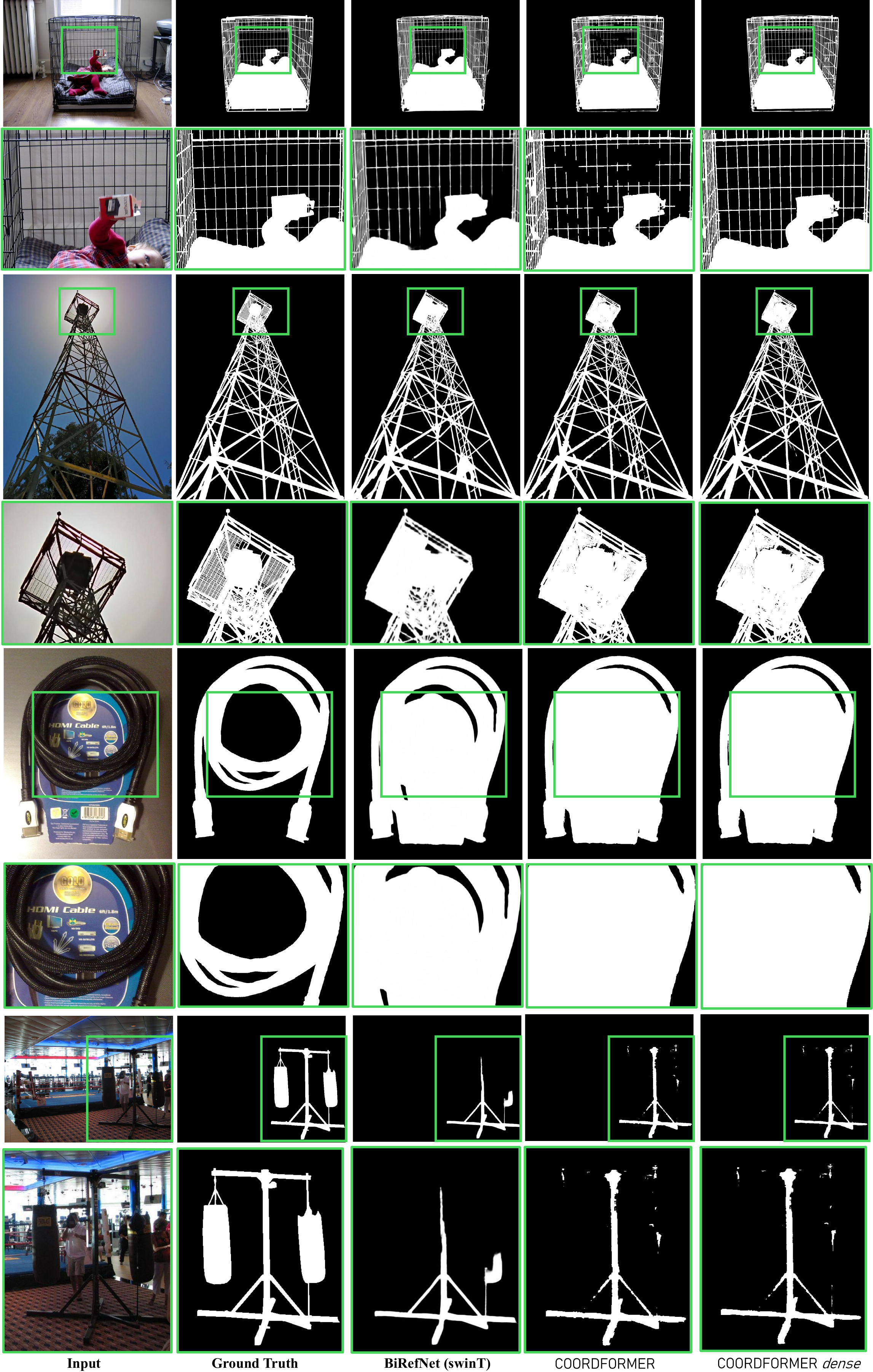}
    \caption{\textbf{Failure cases of \algoname{}. Our method is compared with \algoname{} \textit{dense} and BiRefNet on DIS5K \cite{qin2022dis5k}}. Please zoom in for a clearer view.}
    \label{fig:qualitative_supp_dis_fail}
\end{figure*}

Although \algoname{} achieves strong results on both MaSS13K and DIS5K, a few recurring failure modes can be observed, as shown in \cref{fig:qualitative_supp_mass_fail} and \cref{fig:qualitative_supp_dis_fail}.

\noindent On both datasets, the most challenging examples involve extremely small or distant semantic regions, very thin structures, or boundaries embedded in highly cluttered textures. These cases are illustrated in the first two rows of \cref{fig:qualitative_supp_mass_fail} and \cref{fig:qualitative_supp_dis_fail}. In the first rows, the thin structures can be recovered by disabling SEFI, as shown in the \algoname{} \textit{dense} column, where our method uses the full set of image coordinates to generate the prediction. The second rows show cases in which \algoname{}, \algoname{} \textit{dense}, and MaSSFormer/BiRefNet all fail to predict thin cables, branches, or grid-like textures. These results highlight the challenges posed by such structures in both DIS5K and MaSS13K. The third and fourth rows highlight failures related to semantic ambiguity. In particular, the third row of \cref{fig:qualitative_supp_mass_fail} depicts a scene with strong reflections caused by a glass surface, where \algoname{}, MaSSFormer, and \algoname{} \textit{dense} fail to predict the \texttt{vegetation} class. The fourth row of \cref{fig:qualitative_supp_mass_fail} illustrates one of the most common semantic errors, in which the \texttt{other} class is misclassified due to intrinsic ambiguities associated with this category in the MaSS13K dataset. The last two rows of \cref{fig:qualitative_supp_dis_fail} illustrate cases in which the main object is not correctly identified by the compared methods.
Overall, these examples show that the main limitations of \algoname{} arise in the presence of extremely fine structures and semantically ambiguous scenes. While dense inference can recover part of the missing details in some cases, the remaining errors highlight the intrinsic difficulty of these examples for both efficient and dense prediction settings.

\end{document}